%% file: main.tex
\documentclass[10pt,twocolumn,letterpaper]{article}

\usepackage{cvpr}              

\input{preamble}

\definecolor{cvprblue}{rgb}{0.21,0.49,0.74}
\usepackage[pagebackref,breaklinks,colorlinks,allcolors=cvprblue]{hyperref}

\def\confName{CVPR}
\def\confYear{2027}

\title{HINT-WAM: Hindsight-Informed Training for World Action Models}

\author{\textbf{Phuong Tuan Dat}$^{1}$, \textbf{Qi Li}$^{1}$, \textbf{Xinchao Wang}$^{1,\dagger}$\\
{\normalsize $^{1}$National University of Singapore}\\
{\normalsize \{phuongtuandat, liqi\}@u.nus.edu, xinchao@nus.edu.sg}\\
{\normalsize $^{\dagger}$Corresponding author}
}

\begin{document}
\twocolumn[{%
\renewcommand\twocolumn[1][]{#1}%
\maketitle
\begin{center}
    \centering
    \includegraphics[width=\textwidth,trim=93pt 394pt 76pt 152pt,clip]{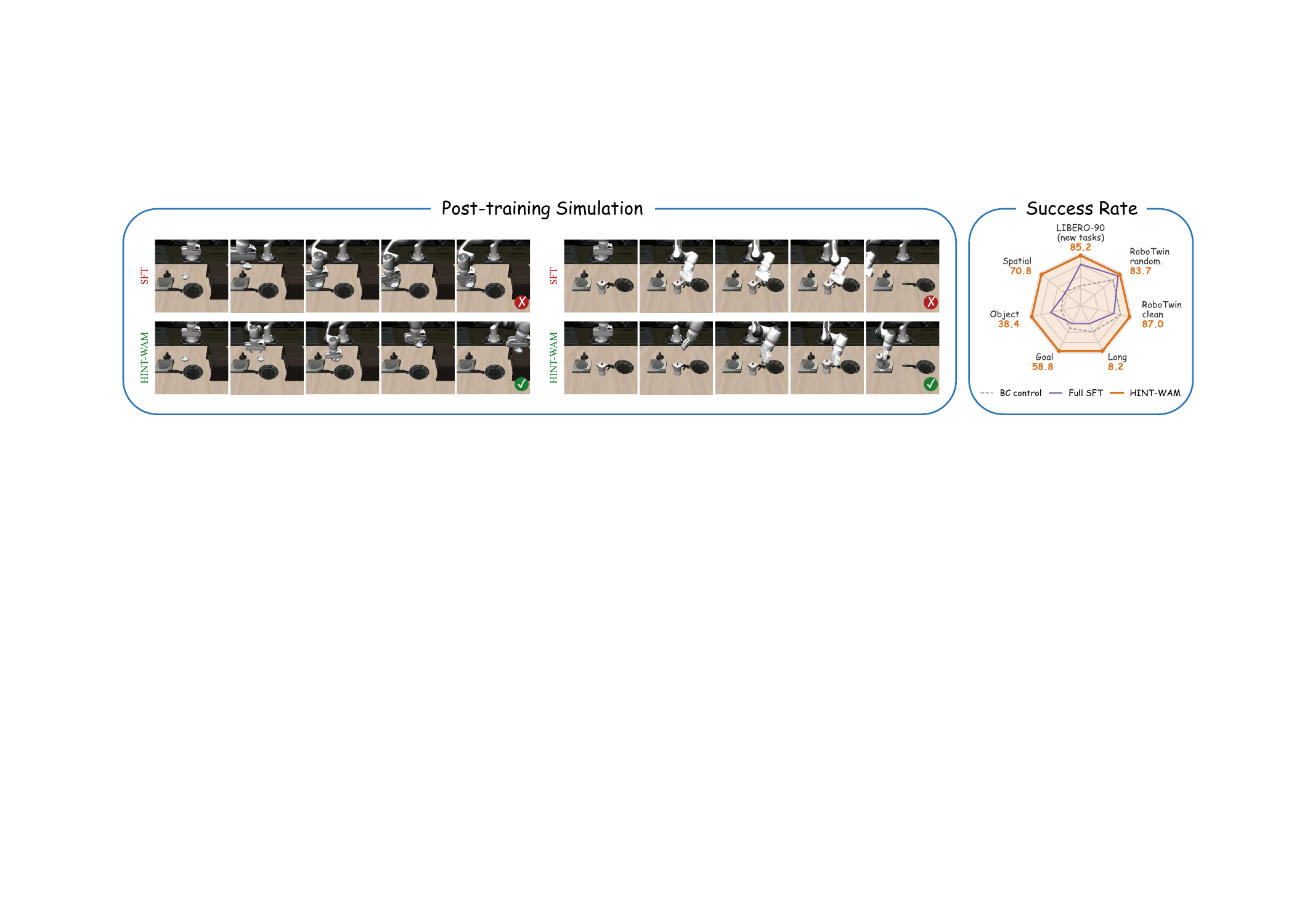}
    \captionof{figure}{\textbf{HINT-WAM post-trains world action models with hindsight.} \emph{Left:} rollouts on two LIBERO-90 tasks from the same initial state, where SFT fails and HINT-WAM succeeds. \emph{Right:} success rates (\%) of HINT-WAM, the BC control, and full SFT on the new LIBERO-90 tasks, RoboTwin 2.0, and the four original LIBERO suites after post-training on LIBERO-90 without replay.}
    \label{fig:teaser}
\end{center}%
\vspace{0.5em}
}]
\input{sec/0_abstract_arxiv}    
\input{sec/1_intro_arxiv}
\input{sec/2_related}
\input{sec/3_method}
\input{sec/4_experiments}
\input{sec/5_conclusion_arxiv}

{
    \small
    \bibliographystyle{ieeenat_fullname}
    \bibliography{main}
}

\input{sec/X_suppl}

\end{document}

%% file: preamble.tex
\usepackage{algorithm}
\usepackage{algpseudocode}
\usepackage{multirow}

%% file: sec/0_abstract_arxiv.tex
\begin{abstract}
World action models (WAMs) have shown promising generalization and data efficiency in robot manipulation by coupling action generation with predictions of future visual dynamics. Translating these capabilities into reliable task execution, however, remains a challenge for existing post-training methods. Offline supervised fine-tuning provides dense supervision on expert trajectories, but its flow-matching action expert is trained only on noised demonstration actions and never on the intermediate states produced by its own sampler; task-specific adaptation can also compromise pretrained capabilities. On-policy reinforcement learning trains on the policy's own behavior, yet sparse task rewards make exploration and credit assignment difficult, while dense reward shaping still provides scalar evaluations rather than direct action targets. We introduce \textbf{HINT-WAM}, the first on-policy self-distillation framework that supervises the action expert on its own sampling trajectories using hindsight, the future video of the demonstration, as privileged information available to the WAM itself. For each demonstration observation, the student generates its action through its own denoising trajectory, and at the states along this trajectory the same model acts as a teacher conditioned additionally on this hindsight, providing a conditional action distribution as a dense training target, while the student sees only deployment-time inputs and needs no environmental rewards or simulator interaction. To transfer this supervision to a flow-matching action expert, HINT-WAM distills the teacher into the student with an energy score, a distribution-level objective that matches the teacher's action distribution from the student's own samples rather than regressing its velocity field. The project web is available in this \href{https://tuandattt.github.io/hintwam/}{link}.
\end{abstract}

%% file: sec/1_intro_arxiv.tex
\section{Introduction}
\label{sec:intro}

World action models (WAMs) have demonstrated promising generalization and data efficiency for robot manipulation by coupling action generation with future visual prediction~\cite{hu2025vpp,zhu2025uwm,li2026lingbotva}. Recent models generate continuous action chunks through flow matching, providing an expressive interface between visual predictions and robot motion~\cite{li2026lingbotva,akbari2026flashwam}. Despite this progress, reliable execution on downstream tasks~\cite{li2026vision,li2026badwam} still requires adaptation to task-specific demands and to the deviations that arise during closed-loop interaction.

Supervised fine-tuning (SFT) offers a straightforward adaptation strategy by providing dense action targets from expert demonstrations. However, SFT supervises the model only on demonstration data. During training, the action expert sees only noised versions of the ground-truth actions and never the intermediate states produced by its own sampler, a form of exposure bias~\cite{ning2023exposure} and the sampler-level counterpart of the compounding errors at deployment~\cite{ross2011dagger}, whose environment-level gap requires closed-loop interaction. Moreover, given only the current observation, a demonstrated action chunk is one of several plausible continuations, and the SFT target does not reveal which one the demonstrator followed. Finally, adapting a pretrained model to a narrow task distribution can degrade previously acquired capabilities~\cite{shenfeld2025rlrazor}.

On-policy reinforcement learning (RL) addresses the state-distribution mismatch by optimizing the policy on its own rollouts, motivating its use in VLA and WAM post-training~\cite{li2025simplevlarl,lu2025vlarl,qian2026wamrl}. However, many manipulation tasks offer only sparse success rewards, making exploration and credit assignment costly when successful trajectories are rarely sampled. Even with denser, reconstruction-based rewards~\cite{qian2026wamrl}, reward-based optimization only evaluates sampled behavior through scalar feedback, leaving the policy to infer how its actions should change. This motivates supplying structured action supervision directly on the states produced by the policy's own action sampler.

Our key insight is that the future frames of a demonstration, which we call \emph{hindsight}, carry privileged information about the intended task progression, connecting learning with privileged information~\cite{vapnik2009lupi,chen2019lbc} to on-policy distillation~\cite{agarwal2024gkd}. We build on this insight in \textbf{HINT-WAM}: the same model serves as a teacher that additionally reads the hindsight, whereas the student observes only the inputs available at deployment and is supervised on the states produced by its own action sampler. Throughout this paper, \emph{on-policy} refers to the generation of actions: as in on-policy distillation for language models~\cite{agarwal2024gkd}, where prompts are drawn from a dataset and the student generates the response, observations are drawn from demonstrations and the student generates the denoising trajectory of its action chunk. We do not collect new environment states.

Transferring this supervision to a flow-matching action expert introduces a further challenge. Such experts generate continuous actions by integrating a velocity field, so the token-level probability matching used in autoregressive distillation does not apply directly. Regressing the teacher's velocity field would drive the student toward the teacher's conditional mean, which averages over modes that the student cannot distinguish without hindsight. We therefore distill the teacher with the energy score~\cite{gneiting2007scoring}, a strictly proper scoring rule that matches the teacher's action distribution from the student's own samples and one teacher sample.
Our contributions are summarized as follows:
\begin{itemize}
    \item We introduce HINT-WAM, a self-distillation framework for world action models in which the policy acts as its own teacher given hindsight, requiring neither a separately trained teacher nor environmental rewards or simulator interaction.
    \item We propose a distribution-level distillation objective for flow-matching action experts based on the energy score, which avoids the mode averaging of velocity regression and requires no action-likelihood evaluation. The framework adapts the action expert while keeping the world-modeling backbone frozen.
    \item Comprehensive experiments on LIBERO-90 and RoboTwin 2.0 demonstrate that HINT-WAM consistently improves post-training performance over SFT and matched baselines, while mitigating the catastrophic forgetting of pretrained skills.
\end{itemize}

%% file: sec/2_related.tex
\section{Related Work}
\label{sec:related}

\subsection{World Action Models}
\label{sec:related_wam}

Vision-language-action (VLA) models combine large-scale multimodal pretraining with diverse robot demonstrations to translate camera observations and language instructions into robot commands~\cite{brohan2023rt1,zitkovich2023rt2,oneill2024openx,kim2024openvla}. Rather than discretizing actions into tokens, many recent policies generate continuous action chunks with diffusion~\cite{chi2023diffusionpolicy} or flow matching~\cite{lipman2023flow,black2024pi0}. World action models (WAMs) additionally reason about how the scene will evolve~\cite{shen2026worldactionmodelssurvey}, and differ mainly in how tightly they tie prediction to control. Some extract predictive visual representations from a video model to condition a separate action head~\cite{hu2025vpp}, while others generate future frames and actions jointly through a shared denoising process or a unified video--action backbone~\cite{guo2024pad,li2025uva,zhu2025uwm}. Autoregressive designs such as LingBot-VA interleave video and action prediction for closed-loop control~\cite{li2026lingbotva}, and RoboDreamer exploits imagined dynamics for planning~\cite{zhou2024robodreamer}. Recent architectures pair a large world-modeling backbone with a lightweight flow-matching action expert~\cite{akbari2026flashwam}. While these works study WAM design and pretraining, HINT-WAM post-trains only the action expert on the states the policy visits.

\subsection{On-Policy Self-Distillation}
\label{sec:related_opsd}

On-policy distillation builds on the observation, first formalized by DAgger~\cite{ross2011dagger}, that a policy learns most effectively from the states it reaches on its own rather than from those recorded by a demonstrator. GKD~\cite{agarwal2024gkd} and MiniLLM~\cite{gu2024minillm} adopt this principle for compressing large language models: the student generates its own sequences, which the teacher then scores token by token, and MiniLLM additionally argues for the mode-seeking reverse KL as the distillation objective. A more recent line of work removes the need for a separate, larger teacher by letting a single model play both roles, where the teacher is the model itself given privileged information that the student does not see. Self-Distilled Reasoner~\cite{zhao2026opsd} trains a reasoning model on its own on-policy samples under the supervision of a copy conditioned on reference solutions, dOPSD~\cite{dat2026dopsd} extends this self-distillation scheme to diffusion language models, and PRISM~\cite{dat2026prism} applies it to the design of communication topologies in multi-agent systems, guiding the designer with a privileged sibling. All of these methods operate on discrete tokens and rely on likelihood-based objectives. HINT-WAM brings on-policy self-distillation to embodied control, where the student is a flow-matching action expert that generates continuous actions without a tractable likelihood. Unlike prior flow-matching distillation methods, which regress the teacher's velocity field~\cite{fang2026flowopd,li2026diffusionopd}, we match action distributions with a proper scoring rule, the energy score~\cite{gneiting2007scoring}, which needs only one teacher sample per state.

\subsection{On-Policy Distillation for World Action Models}
\label{sec:related_wamopd}

Recent works bring on-policy distillation to WAMs, supervising the student on its own rollouts with a separate, stronger task-specific teacher, either through joint video--action targets~\cite{yang2026wamopd} or by replaying a fixed pool of student and teacher trajectories~\cite{liu2026wamopd}. The former queries the teacher for paired video and action targets on student rollout histories, while the latter reweights each replayed decision with prefix-based importance weights to correct the shift between stored and current histories. These methods require an additional teacher model, making post-training computationally expensive. In contrast, HINT-WAM derives the teacher from the student itself by adding only a lightweight hindsight branch. To the best of our knowledge, it is the first self-distillation framework for world action models.

%% file: sec/3_method.tex
\section{Method}
\label{sec:method}

This section presents HINT-WAM, which post-trains a pretrained WAM by letting the model act as its own teacher given hindsight. We first review the base model (\cref{sec:method_prelim}) and outline the framework (\cref{sec:method_overview}), then describe the hindsight teacher (\cref{sec:method_teacher}) and its distillation into the student (\cref{sec:method_opsd}).

\subsection{Preliminaries}
\label{sec:method_prelim}

\paragraph{Base world action model.}
We build on Fast-WAM~\cite{yuan2026fastwam}, a world action model whose backbone is a Mixture-of-Transformers~\cite{liang2024mot} with two experts that share a single attention operation in every layer. The \emph{video expert} is a large pretrained video diffusion transformer~\cite{wan2025wan} operating on video VAE latents, and the \emph{action expert} is a lightweight diffusion transformer that generates an action chunk $\mathbf{a}\in\mathbb{R}^{H\times D}$ of $H$ control steps. The policy is conditioned on an observation $o=(I,\ell,s)$ consisting of the current camera frame $I$, the language instruction $\ell$, and the proprioceptive state $s$. Action tokens attend only to the video tokens of the current frame, so the action expert never observes future frames.

\paragraph{Flow-matching action expert.}
Following the rectified-flow convention~\cite{lipman2023flow,liu2023rectifiedflow}, $\sigma=0$ corresponds to data and $\sigma=1$ to noise. A noisy action chunk is $\mathbf{a}_\sigma=(1-\sigma)\mathbf{a}+\sigma\boldsymbol{\epsilon}$ with $\boldsymbol{\epsilon}\sim\mathcal{N}(\mathbf{0},\mathbf{I})$, and the action expert predicts a velocity $v_\theta(\mathbf{a}_\sigma,\sigma,o)$ trained with
\begin{equation}
    \mathcal{L}_{\mathrm{FM}}(\theta)=\mathbb{E}_{(o,\mathbf{a}),\sigma,\boldsymbol{\epsilon}}\big\|v_\theta(\mathbf{a}_\sigma,\sigma,o)-(\boldsymbol{\epsilon}-\mathbf{a})\big\|_2^2 .
    \label{eq:fm}
\end{equation}
At inference, the policy integrates the probability-flow ODE from $\mathbf{a}_1=\boldsymbol{\epsilon}$ to $\sigma=0$ with $N$ Euler steps on a decreasing schedule $1=\sigma_1>\dots>\sigma_N>\sigma_{N+1}=0$:
\begin{equation}
    \mathbf{a}_{\sigma_{i+1}}=\mathbf{a}_{\sigma_i}+\Delta_i\, v_\theta(\mathbf{a}_{\sigma_i},\sigma_i,o),\qquad \Delta_i=\sigma_{i+1}-\sigma_i<0 .
    \label{eq:euler}
\end{equation}
Since the video expert only processes the current frame, its keys and values are computed once per observation and cached, so every ODE step evaluates only the action expert. Finally, every point on a sampling trajectory yields an estimate of the clean action in a single forward pass,
\begin{equation}
    \hat{\mathbf{a}}_\theta(\mathbf{a}_\sigma)=\mathbf{a}_\sigma-\sigma\, v_\theta(\mathbf{a}_\sigma,\sigma,o),
    \label{eq:implied}
\end{equation}
which we call the \emph{implied action}. For the optimal field it is the conditional mean $\mathbb{E}[\mathbf{a}\mid\mathbf{a}_\sigma]$ rather than the ODE endpoint, but at $\sigma_N$ it coincides with the final Euler step, \ie, with the policy's output.

\subsection{Overview}
\label{sec:method_overview}

\begin{figure*}[t]
    \centering
    \includegraphics[width=\textwidth]{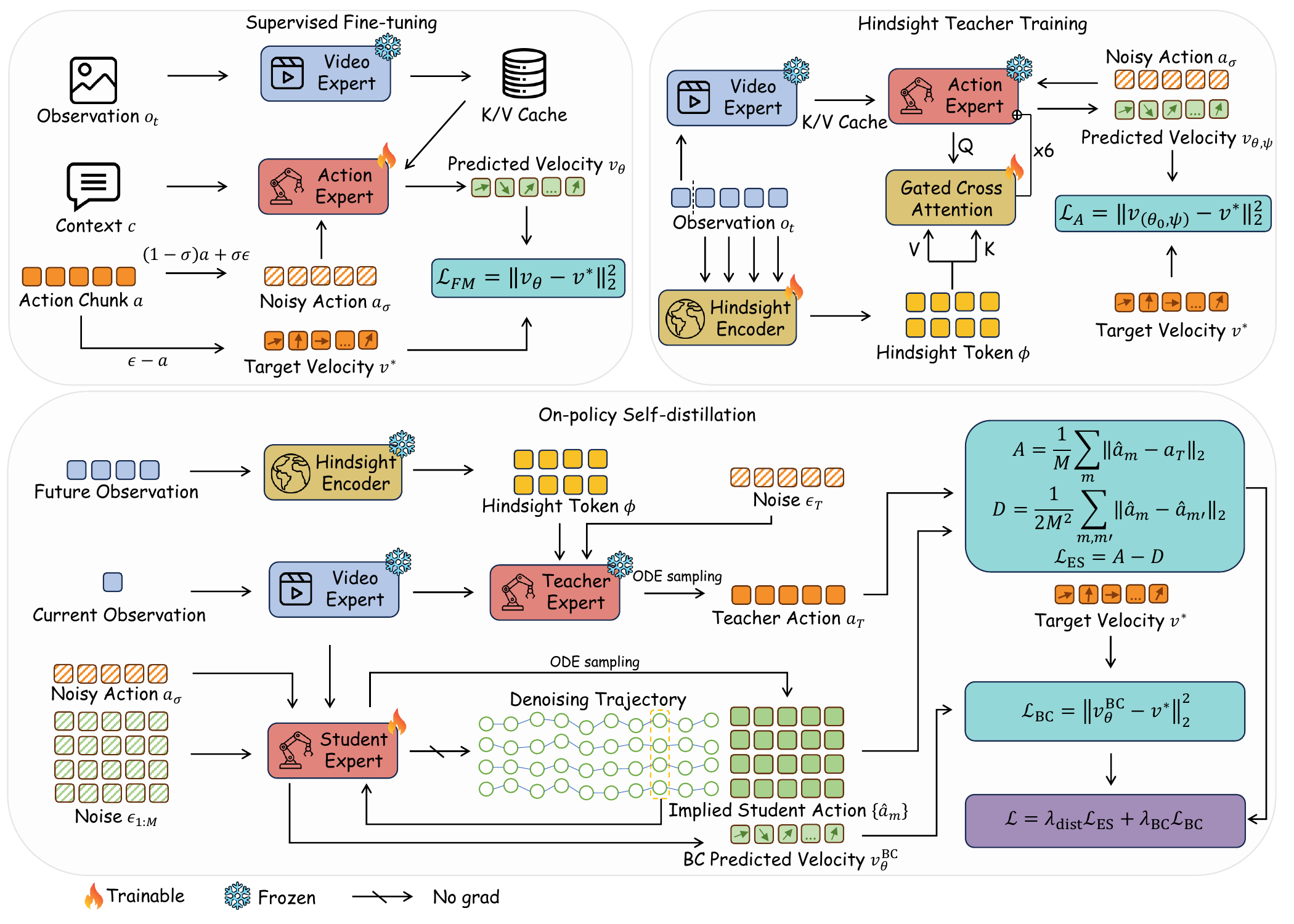}
    \caption{\textbf{Overview of HINT-WAM.}
    \textbf{Stage 0 (top left):} the action expert is fine-tuned with flow matching on demonstrations, conditioned on the key/value cache of the frozen video expert, giving the student initialization.
    \textbf{Stage A (top right):} with the SFT model frozen, a hindsight encoder compresses the demonstration's future observations into hindsight tokens, which the action expert reads through gated cross-attention adapters; only the encoder and adapters are trained.
    \textbf{Stage B (bottom):} the student integrates its own ODE without gradients, and its implied actions at selected denoising points are matched to a single sample from the frozen hindsight teacher with the energy score, together with a behavior-cloning anchor. Only the student is deployed.}
    \label{fig:overview}
\end{figure*}

Standard SFT minimizes \cref{eq:fm} on the demonstrations and therefore suffers from exposure bias and target ambiguity (\cref{sec:intro}). Following on-policy self-distillation for language models~\cite{zhao2026opsd}, we let a single network serve as both teacher and student, where the teacher additionally receives \emph{hindsight}~\cite{andrychowicz2017her}: the future video of the demonstration. Hindsight is available in every demonstration at no extra cost and resolves the ambiguity of the SFT target, while distilling on the student's own sampling trajectories addresses the exposure bias. As shown in \cref{fig:overview}, HINT-WAM proceeds in three stages:
\begin{itemize}[leftmargin=*,itemsep=1pt,topsep=2pt]
    \item \textbf{Stage 0 (SFT).} The action expert is fine-tuned on the demonstrations with \cref{eq:fm}, giving the student initialization $\theta_0$.
    \item \textbf{Stage A (hindsight teacher).} With $\theta_0$ frozen, a lightweight hindsight encoder and gated adapters learn to read the future video, giving the teacher $T=(\theta_0,\psi)$.
    \item \textbf{Stage B (on-policy self-distillation).} The student $\theta$, initialized from $\theta_0$, is trained to match the teacher on its own sampling trajectories, with only the action expert updated.
\end{itemize}
The video expert stays frozen in every stage. At deployment, only the student is used, so the policy never requires access to the future.

\subsection{Hindsight Teacher}
\label{sec:method_teacher}

The teacher is the student network $\theta_0$ equipped with two small trainable modules: a hindsight encoder that summarizes the future of the demonstration, and gated cross-attention adapters that allow the frozen action expert to read this summary.

\paragraph{Hindsight encoder.}
Each training sample is a demonstration clip whose video window is encoded by the VAE into latents $\mathbf{z}\in\mathbb{R}^{C\times T\times h\times w}$. The first latent frame encodes the current observation, which the student already sees, while the remaining frames describe what happens next in the demonstration over the same horizon as the action chunk. The encoder drops the first latent frame so that the teacher gains no advantage from information shared with the student, mean-pools every remaining frame over space, and projects the resulting sequence with a linear layer followed by LayerNorm. A set of $N_\phi$ learned queries then pools this sequence through multi-head attention~\cite{lee2019settransformer}:
\begin{equation}
\begin{aligned}
    \mathbf{u}&=\mathrm{LN}\big(W_p\,\mathrm{pool}_{hw}(\mathbf{z}_{1:})\big),\\
    \phi_\psi(\mathbf{z})&=\mathrm{LN}\big(W_o\,\mathrm{MHA}(Q,\mathbf{u})\big)\in\mathbb{R}^{N_\phi\times d_\phi},
\end{aligned}
\label{eq:hindsight}
\end{equation}
where $\mathbf{u}$ is the projected sequence of future frames and $Q$ denotes the learned queries. Spatial pooling discards the scene layout, already visible in the current frame, and keeps the temporal axis that carries the trajectory's future.

\paragraph{Gated cross-attention adapters.}
We attach adapters to the action expert after a subset of layers $\mathcal{S}$. For the action-token hidden states $\mathbf{h}$ leaving layer $l\in\mathcal{S}$, the adapter computes
\begin{equation}
\begin{gathered}
    \mathbf{q}=W_{\mathrm{down}}\mathrm{LN}(\mathbf{h}),\quad \mathbf{k}=W_k\phi,\quad \mathbf{v}=W_v\phi,\\
    \mathbf{h}'=\mathbf{h}+\tanh(\alpha_l)\, W_{\mathrm{up}}W_{\mathrm{out}}\,\mathrm{Attn}(\mathbf{q},\mathbf{k},\mathbf{v}),
\end{gathered}
\label{eq:adapter}
\end{equation}
with a tanh gate in the style of Flamingo~\cite{alayrac2022flamingo}. The adapter runs beside the frozen block rather than inside the shared attention, so that a zero branch is an exact identity, and we initialize $W_{\mathrm{up}}=\mathbf{0}$ and $\alpha_l=1$, so that the teacher is identical to $\theta_0$ at the first step while the branch still receives gradients~\cite{zhang2023controlnet,peebles2023dit}. Without $\phi$, the adapters are skipped, so student and teacher share all backbone weights.

\paragraph{Training objective.}
The entire backbone $\theta_0$, including both experts, is frozen, and only the encoder and adapter parameters, collectively denoted $\psi$, are trained with the flow-matching loss of \cref{eq:fm} given hindsight:
\begin{equation}
    \mathcal{L}_{A}(\psi)=\mathbb{E}\big\|v_{\theta_0,\psi}\big(\mathbf{a}_\sigma,\sigma,o,\phi_\psi(\mathbf{z})\big)-(\boldsymbol{\epsilon}-\mathbf{a})\big\|_2^2 .
    \label{eq:teacher}
\end{equation}
Freezing $\theta_0$ guarantees that any improvement of the teacher stems from the hindsight $\phi$ rather than from additional fitting capacity. The teacher must be built on the same weights from which the student starts; otherwise it would pull the student toward a different backbone.

\subsection{Distilling Hindsight on Student Trajectories}
\label{sec:method_opsd}

In Stage B, the student $\theta$ is initialized from $\theta_0$ and only its action expert is trained, while the teacher $T=(\theta_0,\psi)$ is kept frozen. Each training sample provides the current observation $o$, the demonstrated chunk $\mathbf{a}$, and the future latents $\mathbf{z}$, seen only by the teacher through $\phi=\phi_\psi(\mathbf{z})$.

\paragraph{On-policy states from the student's sampler.}
Starting from fresh noise, the student integrates its own ODE in \cref{eq:euler} with the deployed schedule and without gradients, and records each visited state $\mathbf{a}_{\sigma_i}$. For the given observation, these are precisely the denoising states that the policy visits at inference. This rollout is on-policy with respect to the action sampler and needs no simulator; the observation itself is taken from the demonstration.

\paragraph{Where to supervise.}
The local error of an Euler step grows with $\Delta_i^2$, so we weight trajectory points by $w_i\propto\Delta_i^2$. On the shifted schedule we use, this weighting concentrates most of its mass on the last few points. Rather than sampling $\sigma$, we therefore enumerate the top-$K$ points and renormalize their weights to $\bar{w}_k$. Gradients enter only by re-evaluating the student at these points, and we never backpropagate through the sequential rollout.

\paragraph{Velocity matching.}
A direct objective matches the teacher's velocity at the selected states. With $v^k_\theta=v_\theta(\mathbf{a}_{\sigma_k},\sigma_k,o)$ and $v^k_T=v_T(\mathbf{a}_{\sigma_k},\sigma_k,o,\phi)$, we minimize
\begin{equation}
    \mathcal{L}_{\mathrm{vel}}(\theta)=\sum_{k=1}^{K}\bar{w}_k\,\mathbb{E}\big\|v^k_\theta-\mathrm{sg}\big[v^k_T\big]\big\|_2^2 ,
    \label{eq:vel}
\end{equation}
where $\mathrm{sg}[\cdot]$ is the stop-gradient operator and padded steps of the chunk are masked out.

\paragraph{Distribution matching with the energy score.}
For a fixed student input, the minimizer of \cref{eq:vel} is the teacher's conditional mean given what the student observes. Because the student cannot see $\phi$, the teacher's behavior is multi-modal from the student's perspective, and its mean may be an action that is correct under none of the modes. We therefore match \emph{distributions} with the energy score~\cite{gneiting2007scoring,szekely2013energy}, a strictly proper scoring rule that requires only a single sample from the target. Concretely, the student draws $M$ independent trajectories and the teacher draws one action $\mathbf{a}_T$ by integrating its own ODE with $\phi$ from fresh noise, both without gradients. At the selected point $\sigma_k$ of the $m$-th student trajectory, a differentiable student sample is obtained in a single forward pass through the implied action of \cref{eq:implied}, $\hat{\mathbf{a}}_{m,k}=\mathbf{a}^m_{\sigma_k}-\sigma_k\,v_\theta(\mathbf{a}^m_{\sigma_k},\sigma_k,o)$. The energy score over whole action chunks is
\begin{equation}
\begin{aligned}
    \mathcal{L}_{\mathrm{ES}}(\theta)=&\sum_{k=1}^{K}\bar{w}_k\Big[\frac{1}{M}\sum_{m=1}^{M}\big\|\hat{\mathbf{a}}_{m,k}-\mathbf{a}_T\big\|_2\\
    &-\frac{1}{2M^2}\sum_{m,m'=1}^{M}\big\|\hat{\mathbf{a}}_{m,k}-\hat{\mathbf{a}}_{m',k}\big\|_2\Big],
\end{aligned}
\label{eq:es}
\end{equation}
where the norm is taken over the flattened $H\times D$ chunk. The first term pulls the student's samples toward the teacher's sample, and the second term pushes them apart, which prevents collapse onto the conditional mean. The population energy score is strictly proper, minimized only when the student's distribution equals the teacher's. \cref{eq:es} is its plug-in estimator, whose expectation weights the repulsion by $(M-1)/M$ (the unbiased estimator uses $1/[2M(M-1)]$ over $m\neq m'$); it thus slightly under-weights dispersion, a bias vanishing as $M$ grows.

\paragraph{Behavior-cloning anchor.}
Pure distillation would also follow the teacher where the teacher is wrong. We bound this effect with a flow-matching term on the demonstrations, computed for the student without hindsight, $\mathcal{L}_{\mathrm{BC}}(\theta)=\mathcal{L}_{\mathrm{FM}}(\theta)$. The Stage-B objective is:
\begin{equation}
    \mathcal{L}_{B}(\theta)=\lambda_{\mathrm{dist}}\,\mathcal{L}_{\mathrm{dist}}(\theta)+\lambda_{\mathrm{BC}}\,\mathcal{L}_{\mathrm{BC}}(\theta),
    \label{eq:total}
\end{equation}
where $\mathcal{L}_{\mathrm{dist}}\in\{\mathcal{L}_{\mathrm{vel}},\mathcal{L}_{\mathrm{ES}}\}$ is the distillation objective, and $\lambda_{\mathrm{dist}}$ and $\lambda_{\mathrm{BC}}$ balance the two terms. \cref{alg:opsd} summarizes one training step.

\begin{algorithm}[t]
\caption{One training step of HINT-WAM (Stage B).}
\label{alg:opsd}
\begin{algorithmic}[1]
\Require batch $\{(o,\mathbf{a},\mathbf{z})\}$, student $\theta$, frozen teacher $T=(\theta_0,\psi)$, $K$, $M$, $N$, $\lambda_{\mathrm{dist}}$, $\lambda_{\mathrm{BC}}$
\State $\mathcal{C}\gets$ prefill video keys/values from the current frame in $o$ \Comment{shared below}
\State $\phi\gets\phi_\psi(\mathbf{z})$ \Comment{hindsight, teacher only}
\For{$m=1,\dots,M$}
    \State $\{\mathbf{a}^m_{\sigma_i}\}_{i=1}^{N}\gets$ \Call{Rollout}{$v_\theta(\cdot,\cdot,o;\mathcal{C}),\boldsymbol{\epsilon}_m$} \Comment{no grad}
\EndFor
\State $\{(\sigma_k,\bar{w}_k)\}_{k=1}^{K}\gets$ top-$K$ points by $\Delta_i^2$
\State $\mathbf{a}_T\gets$ \Call{Rollout}{$v_T(\cdot,\cdot,o,\phi;\mathcal{C}),\boldsymbol{\epsilon}_T$}.final \Comment{no grad}
\State $\hat{\mathbf{a}}_{m,k}\gets\mathbf{a}^m_{\sigma_k}-\sigma_k\,v_\theta(\mathbf{a}^m_{\sigma_k},\sigma_k,o;\mathcal{C})$ \Comment{with grad}
\State $\mathcal{L}_{\mathrm{dist}}\gets$ energy score of $\{\hat{\mathbf{a}}_{m,k}\}$ against $\mathbf{a}_T$ \Comment{\cref{eq:es}}
\State $\mathcal{L}_{\mathrm{BC}}\gets$ flow-matching loss of $\theta$ on $(o,\mathbf{a})$ \Comment{\cref{eq:fm}}
\State update the action expert of $\theta$ with $\nabla_\theta\big[\lambda_{\mathrm{dist}}\mathcal{L}_{\mathrm{dist}}+\lambda_{\mathrm{BC}}\mathcal{L}_{\mathrm{BC}}\big]$
\end{algorithmic}
\end{algorithm}

%% file: sec/4_experiments.tex
\section{Experiments}
\label{sec:experiments}

\subsection{Implementation}
\label{sec:exp_impl}

\paragraph{Datasets.}
We evaluate HINT-WAM on two simulation benchmarks, LIBERO~\cite{liu2023libero} and RoboTwin 2.0~\cite{chen2025robotwin2}. Our main benchmark is LIBERO-90, whose 90 tasks share the robot and action space of the four standard LIBERO suites on which the released Fast-WAM checkpoint was trained, yet are unseen by it. We post-train on its 3{,}921 demonstrations and evaluate each task with 50 trials from the fixed initial states of LIBERO. To measure forgetting, we additionally evaluate the 40 tasks of the four original suites, which also provide a small replay memory for our continual variant. RoboTwin 2.0 covers dual-arm manipulation over 50 tasks. Since the released checkpoint was trained on these tasks, we evaluate on a hard subset of 25 tasks, selected before any training, with 50 episodes each in the clean and randomized settings. All methods share the same initial states, number of trials, and sampler.

\paragraph{Training details.}
We keep the architecture and sampler of the released Fast-WAM checkpoints unchanged, with 32-step action chunks generated by 10 Euler steps. The hindsight encoder and the adapters add about 13M parameters ($0.2\%$), used only by the teacher. All stages use AdamW~\cite{loshchilov2019adamw} with a cosine schedule and an effective batch size of 192 on 8 A5000 GPUs. Stage 0 fine-tunes the action expert for 5{,}000 steps (learning rate $10^{-4}$), Stage A trains the teacher modules for 2{,}000 steps ($3\times10^{-4}$), and Stage B trains the action expert for 2{,}000 steps ($10^{-4}$) with $M=4$, $K=1$, and $\lambda_{\mathrm{dist}}=\lambda_{\mathrm{BC}}=1$. Stages A and B start from SFT step 3{,}000, while we report the final step 5{,}000 as the SFT baseline. On RoboTwin, the released checkpoint serves as the SFT model, and all hyperparameters are reused without tuning. For continual post-training, the replay memory~\cite{chaudhry2019tiny} is mixed into every stage, so that the teacher also learns to use hindsight on the old tasks. Further details are given in the supplementary material.

\subsection{Comparison Results}
\label{sec:exp_main}

We compare HINT-WAM with the following baselines, which share its starting checkpoint, data, trainable parameters, optimizer, and training budget unless stated otherwise:
\begin{itemize}[leftmargin=*,itemsep=1pt,topsep=2pt]
    \item \textbf{Fast-WAM}~\cite{yuan2026fastwam}: the released checkpoint.
    \item \textbf{SFT}: fine-tunes the action expert with \cref{eq:fm}, without a teacher or on-policy states.
    \item \textbf{Full SFT}: fine-tunes both the video and action experts.
    \item \textbf{BC control}: continues SFT with the BC loss only ($\lambda_{\mathrm{dist}}=0$), testing the effect of additional training.
    \item \textbf{Self-imitation}: adds successful rollouts of the SFT policy from randomized placements to the demonstrations and trains with behavior cloning.
    \item \textbf{Shuffled hindsight}: runs HINT-WAM, but the teacher reads the future of \emph{another} episode, removing only the privileged information.
\end{itemize}
On RoboTwin, we compare with Fast-WAM, full SFT, and the BC control.

\begin{table}[t]
    \centering
    \small
    \setlength{\tabcolsep}{5pt}
    \caption{\textbf{Post-training performance on LIBERO-90 and RoboTwin 2.0.} All values are success rates (\%). --: not applicable or not evaluated.}
    \label{tab:main}
    \begin{tabular}{@{}lccc@{}}
        \toprule
        \multirow{2}{*}{Method} & \multirow{2}{*}{LIBERO-90} & \multicolumn{2}{c}{RoboTwin 2.0} \\
        \cmidrule(l){3-4}
        & & Clean & Random. \\
        \midrule
        Fast-WAM~\cite{yuan2026fastwam} & 15.56 & 86.48 & 83.44 \\
        SFT                             & 78.88 & -- & -- \\
        Full SFT                        & 81.24 & 84.12 & 83.24 \\
        BC control                      & 71.56 & 85.36 & 82.08 \\
        Self-imitation                  & 79.62 & -- & -- \\
        Shuffled hindsight              & 80.28 & -- & -- \\
        \midrule
        \textbf{HINT-WAM (Ours)}        & \textbf{85.18} & \textbf{87.04} & \textbf{83.68} \\
        \bottomrule
    \end{tabular}
\end{table}

\begin{table*}[t]
    \centering
    \footnotesize
    \setlength{\tabcolsep}{9pt}
    \renewcommand{\arraystretch}{0.92}
    \caption{\textbf{Continual learning on LIBERO.} Success rates (\%) on the new LIBERO-90 tasks and the four original suites after post-training on LIBERO-90, without and with experience replay.}
    \label{tab:cl}
    \begin{tabular}{@{}lcccccc@{}}
        \toprule
        \multirow{2}{*}{Method} & New tasks & \multicolumn{5}{c}{Original suites} \\
        \cmidrule(lr){2-2}\cmidrule(l){3-7}
        & LIBERO-90 & Spatial & Object & Goal & Long & Avg \\
        \midrule
        Fast-WAM~\cite{yuan2026fastwam} & 15.56 & 96.8 & 99.6 & 95.8 & 94.0 & 96.55 \\
        \midrule
        \multicolumn{7}{@{}l}{\emph{Without replay}} \\
        SFT                & 78.88 & 58.6 & 33.2 & 53.4 & 4.6 & 37.45 \\
        Full SFT           & 81.24 & 54.4 & 30.2 & 44.6 & 2.2 & 32.85 \\
        BC control         & 71.56 & 55.4 & 25.6 & 47.0 & 4.0 & 33.00 \\
        Self-imitation     & 79.62 & 64.4 & \textbf{39.8} & 54.8 & 2.4 & 40.35 \\
        Shuffled hindsight & 80.28 & 66.2 & 36.0 & \textbf{60.2} & 2.4 & 41.20 \\
        \textbf{HINT-WAM (Ours)} & \textbf{85.18} & \textbf{70.8} & 38.4 & 58.8 & \textbf{8.2} & \textbf{44.05} \\
        \midrule
        \multicolumn{7}{@{}l}{\emph{With replay}} \\
        SFT                & 76.70 & 97.2 & 95.8 & 92.0 & 81.2 & 91.55 \\
        Full SFT           & 77.64 & 97.8 & \textbf{96.0} & 94.4 & 83.4 & 92.90 \\
        \textbf{HINT-WAM (Ours)} & \textbf{81.56} & \textbf{98.4} & \textbf{96.0} & \textbf{95.8} & \textbf{90.2} & \textbf{95.10} \\
        \bottomrule
    \end{tabular}
\end{table*}

\begin{figure*}[t]
    \centering
    \setlength{\abovecaptionskip}{3pt}
    \begin{minipage}[t]{0.48\textwidth}
        \centering
        \includegraphics[width=\linewidth]{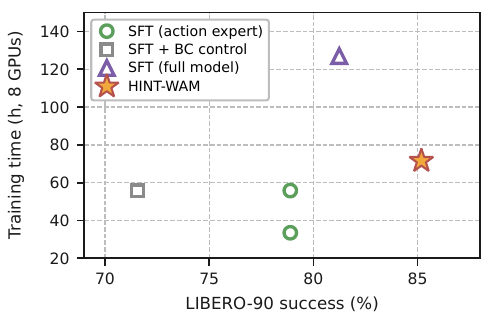}
        \caption{\textbf{Success vs. training time on LIBERO-90.} HINT-WAM reaches the highest success at a fraction of the cost of full SFT, while further SFT or BC training brings no gain.}
        \label{fig:compute}
    \end{minipage}\hfill
    \begin{minipage}[t]{0.48\textwidth}
        \centering
        \includegraphics[width=\linewidth]{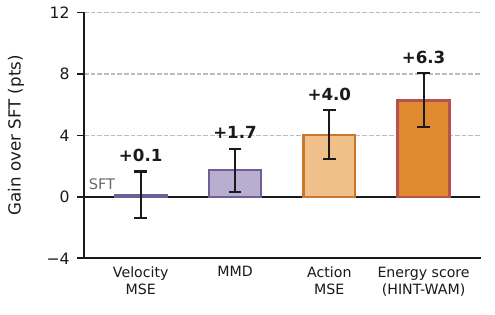}
        \caption{\textbf{Ablation on the distillation objective.} Gain in LIBERO-90 success over SFT with velocity matching (MSE), kernel MMD, action-space MSE, or the energy score.}
        \label{fig:ablation}
    \end{minipage}
\end{figure*}

\begin{figure*}[t]
    \centering
    \setlength{\abovecaptionskip}{3pt}
    \begin{minipage}[t]{0.32\textwidth}
        \centering
        \includegraphics[width=\linewidth]{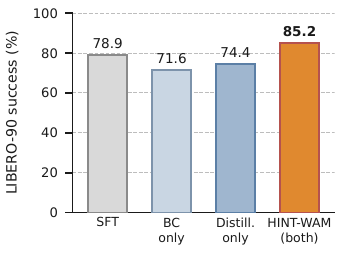}
    \end{minipage}\hfill
    \begin{minipage}[t]{0.32\textwidth}
        \centering
        \includegraphics[width=\linewidth]{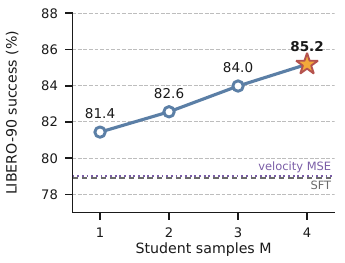}
    \end{minipage}\hfill
    \begin{minipage}[t]{0.32\textwidth}
        \centering
        \includegraphics[width=\linewidth]{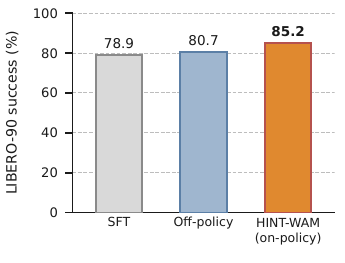}
    \end{minipage}
    \par\vspace{-2pt}
    \begin{minipage}[t]{0.32\textwidth}
        \caption{\textbf{Effect of the BC anchor on LIBERO-90.}}
        \label{fig:anchor}
    \end{minipage}\hfill
    \begin{minipage}[t]{0.32\textwidth}
        \caption{\textbf{Effect of the number of student samples $M$ on LIBERO-90.}}
        \label{fig:num_samples}
    \end{minipage}\hfill
    \begin{minipage}[t]{0.32\textwidth}
        \caption{\textbf{Effect of on-policy states on LIBERO-90.}}
        \label{fig:onpolicy}
    \end{minipage}
\end{figure*}

\paragraph{Results on LIBERO-90.}
As shown in \cref{tab:main}, the released Fast-WAM checkpoint solves only a small fraction of the unseen LIBERO-90 tasks, and SFT raises its success rate to $78.88\%$. HINT-WAM further improves it to $85.18\%$, a gain of $6.30\%$ over SFT on average across the 90 tasks (paired $t=3.58$). It also surpasses full SFT, which fine-tunes both experts, by $3.94\%$ while training only the action expert. The comparison baselines show where this gain comes from. First, it does not stem from additional training: continuing SFT with the same number of steps but without the teacher lowers the success rate to $71.56\%$. Second, it does not stem from training on the policy's own data alone: adding the policy's successful rollouts improves SFT by only $0.74\%$, and HINT-WAM outperforms self-imitation by $5.56\%$ (paired $t=2.96$). Third, the gain is carried by the privileged information: when the teacher reads the future of another episode, the identical pipeline reaches only $80.28\%$, $4.90\%$ below HINT-WAM (paired $t=2.76$).

\paragraph{Results on RoboTwin 2.0.}
On RoboTwin 2.0, the released checkpoint was already trained on the same demonstrations, which leaves little headroom. Continued behavior cloning lowers success in both the clean and the randomized settings, by $1.12\%$ and $1.36\%$, and full SFT also falls below the released checkpoint. In contrast, HINT-WAM stays at or slightly above the released checkpoint and outperforms the BC control by $1.68\%$ in the clean setting (paired $t=2.00$) and by $1.60\%$ in the randomized setting. The gains are smaller than on LIBERO-90 but show the same trend: given the same training budget, hindsight distillation improves the policy where additional imitation degrades it.

\subsection{Continual Learning Results}
\label{sec:exp_cl}

\paragraph{Without replay.}
As shown in \cref{tab:cl}, post-training on LIBERO-90 alone causes severe forgetting on the four original suites for every method, reducing their average success from $96.55\%$ to below $45\%$. HINT-WAM retains the most ($44.05\%$, vs.\ $37.45\%$ for SFT), while full SFT forgets the most ($32.85\%$): updating the video expert as well helps on the new tasks but erodes considerably more of the pretrained skills. Distilled on its own sampling trajectories, the student stays closer to its current behavior than under off-policy imitation, reducing but not preventing forgetting.

\paragraph{With replay.}
Experience replay largely closes the gap, and HINT-WAM composes with it. With replay, HINT-WAM outperforms SFT by $4.86\%$ on the new tasks (paired $t=2.66$) and by $3.55\%$ on the original suites, coming within $1.5\%$ of the released checkpoint, with the largest gain on the long-horizon suite ($9.0\%$). It also leads full SFT by $3.92\%$ and $2.20\%$, respectively, showing that hindsight distillation is complementary to replay.

\subsection{Computational Cost}
\label{sec:exp_cost}

\cref{fig:compute} separates the effect of HINT-WAM from that of additional training. SFT has converged: training from 3{,}000 to 5{,}000 steps costs 22 more hours and leaves success unchanged, and spending the same steps on BC lowers success, suggesting that further imitation overfits. HINT-WAM spends 4.4 hours on the hindsight teacher and 34 hours on distillation, $1.28\times$ the training time of SFT, only $0.56\times$ the 127 hours of full SFT ($2.3\times$ SFT), which still reaches lower success. A distillation step is $1.5\times$ slower than a BC step, a cost paid only during training.

\subsection{Ablation Study}
\label{sec:exp_ablation}

\paragraph{Distribution matching.}
\cref{fig:ablation} compares four distillation objectives under the same teacher, on-policy states, anchor, $M$, and training budget. Velocity matching reaches $79.02\%$, essentially SFT ($+0.14\%$, paired $t=0.09$), as regressing the teacher's velocity drives the student toward its conditional mean. Kernel Maximum Mean Discrepancy (MMD) between the student samples and the teacher sample reaches $80.63\%$: although it matches distributions in principle, its Gaussian-kernel gradients vanish once samples lie farther apart than the bandwidth, as is common for high-dimensional action chunks. Action-space MSE on the implied actions reaches $82.93\%$, but it is minimized by collapsing every student sample onto the teacher's mean, which discards the teacher's uncertainty and conflicts with the flow-matching anchor; its gradient also grows with distance, so a single noisy teacher sample can dominate. The energy score, which is MMD with the distance kernel, avoids both issues: it matches the full distribution, needs no bandwidth, and has unit-norm gradients at every distance. It reaches $85.18\%$, $6.16\%$ above velocity matching (paired $t=3.62$, 48 wins vs.\ 17 losses, sign-test $p=0.0004$).

\paragraph{Behavior-cloning anchor.}
\cref{fig:anchor} shows that neither ingredient helps on its own. Continued imitation lowers success from $78.9\%$ to $71.6\%$, and distillation without the anchor falls to $74.4\%$, below its own SFT initialization, since nothing ties the student to the demonstrations and errors of the teacher and of its own sampling can accumulate. Combined, the two reach $85.2\%$: while they change success by $-7.3\%$ and $-4.5\%$ alone, together they add $+6.3\%$, a strong interaction rather than two additive effects. The anchor thus keeps the student on the demonstration distribution, while distillation corrects it on the states it visits.

\paragraph{Number of student samples.}
\cref{fig:num_samples} varies the number of student samples $M$ in \cref{eq:es}. With $M=1$, the repulsion term vanishes, yet the variant still reaches $81.4\%$ and outperforms velocity matching. Adding samples improves success monotonically to $85.2\%$ at $M=4$, by $1.1$--$1.4\%$ per sample. Action-space MSE with $M=4$, which also lacks repulsion, reaches only $82.9\%$, so the gain at $M=4$ stems from the repulsion that preserves the diversity of the student's samples rather than from the number of samples alone. Since the curve has not saturated, larger $M$ may help further, at one more gradient-free rollout per sample.

\paragraph{On-policy states.}
\cref{fig:onpolicy} isolates where the student is supervised. Replacing the student's sampler states with demonstrated actions noised to the same points $\sigma_k$, \ie, the states seen by SFT, reaches only $80.7\%$, whereas the student's own sampling trajectories reach $85.2\%$. The shuffled-hindsight baseline ($80.3\%$, \cref{tab:main}), which keeps on-policy states but removes the privileged information, shows the converse. Hindsight and on-policy states thus each improve SFT by less than $2\%$ alone but add $6.3\%$ together: the teacher's correction is most useful on the denoising states the student actually reaches at inference.

%% file: sec/5_conclusion_arxiv.tex
\section{Conclusion}
\label{sec:conclusion}

We presented HINT-WAM, the first on-policy self-distillation framework for world action models, in which a hindsight teacher sharing the student's backbone supervises the action expert on its own sampler states with the energy score. Without an external teacher, rewards, or a simulator, HINT-WAM outperforms SFT on LIBERO-90 and RoboTwin 2.0, forgets less, and composes with replay. Since its on-policy states come from the sampler rather than closed-loop interaction, closed-loop data collection is a natural next step.

%% file: sec/X_suppl.tex
\clearpage
\setcounter{page}{1}
\maketitlesupplementary
\appendix

This supplementary material provides additional details and results for HINT-WAM. \cref{sec:supp_impl} describes the implementation and hyperparameters, \cref{sec:supp_cost} the training cost, \cref{sec:supp_teacher} the checks of the hindsight teacher, \cref{sec:supp_stats} the full paired statistics on LIBERO-90, \cref{sec:supp_pertask} a per-task analysis, \cref{sec:supp_robotwin} further RoboTwin 2.0 results, \cref{sec:supp_protocol} the LIBERO-90 evaluation protocol, \cref{sec:supp_selfimit} the self-imitation baseline, \cref{sec:supp_selection} the selection of the objective and learning rate, \cref{sec:supp_toy} a toy experiment on mode averaging, \cref{sec:supp_vis} qualitative rollouts, and \cref{sec:supp_discussion} a discussion of the main design questions.

\section{Implementation Details and Hyperparameters}
\label{sec:supp_impl}

\paragraph{Datasets.}
For LIBERO-90, we use the LeRobot-format release of the 90 tasks, comprising 3{,}921 demonstrations and 569K frames recorded at 20\,Hz. The policy observes an agent-view and a wrist camera at $224\times224$, concatenated horizontally into a single frame, and predicts 7-D actions (end-effector pose deltas and gripper). Forgetting is measured on the 40 tasks of the four original LIBERO suites (Spatial, Object, Goal, and Long) with 50 trials each. For the continual variant, the replay memory contains 5 demonstrations per original task (200 episodes) sampled once with a fixed seed; repeated four times, it accounts for about $19\%$ of the training frames and is mixed into every training stage. On RoboTwin 2.0, the policy observes a head camera and two wrist cameras at $384\times320$ and predicts 14-D actions (joint positions and grippers of both arms). We train on the demonstrations of the 50 tasks used by the released checkpoint and hold out $1\%$ of the episodes. All methods use the normalization statistics of the released checkpoints.

\paragraph{Architecture.}
The Mixture-of-Transformers backbone of Fast-WAM has 30 layers. Its video expert is the Wan2.2-TI2V-5B diffusion transformer (5.00B parameters), and its action expert is a diffusion transformer with 1.02B parameters. Each action chunk spans $H=32$ control steps, and the video window covers 33 frames, of which every fourth frame is encoded by the VAE. Actions are generated with 10 Euler steps on a schedule with shift 5, both during training and at inference.
Our modules are used only by the teacher. The hindsight encoder takes the VAE latents of the future frames, dropping the first latent frame, mean-pools them over space, and applies a linear layer ($48\to512$) followed by LayerNorm. Eight learned queries then pool the sequence with 8-head attention, followed by a linear layer and LayerNorm, so that $\phi\in\mathbb{R}^{8\times512}$. The gated adapters are attached after layers $\{4,9,14,19,24,29\}$ of the action expert with a bottleneck width of 512 and 8 heads, and are initialized with $\alpha_l=1$ and $W_{\mathrm{up}}=\mathbf{0}$, so that the teacher equals the student at initialization. Together, the two modules contain about 13M parameters.

\paragraph{Optimization.}
All stages use AdamW~\cite{loshchilov2019adamw} with $\beta=(0.9,0.95)$, a weight decay of $0.01$, a cosine schedule with $5\%$ linear warm-up, and gradient clipping at $1.0$. Training runs in bfloat16 with DeepSpeed ZeRO-2~\cite{rajbhandari2020zero} on 8 GPUs with 24\,GB of memory each, using seed 42. The video expert is frozen in every stage of every method except full SFT. \cref{tab:train_details} lists the stage-specific settings. In Stage 0, we use $\lambda_{\mathrm{BC}}=2$, which matches the total gradient scale of the distillation stage ($\lambda_{\mathrm{dist}}+\lambda_{\mathrm{BC}}=2$). In Stage B, the energy score uses $M=4$ student samples, $K=1$ supervised point, and one teacher sample, with $\lambda_{\mathrm{dist}}=\lambda_{\mathrm{BC}}=1$. At inference, only the student is used, conditioned on the current frame, the instruction, and the proprioceptive state; on LIBERO, it replans every 10 control steps. On RoboTwin, all hyperparameters are reused from LIBERO without any tuning.

\paragraph{Baselines.}
The BC control and full SFT use the same data, optimizer, schedule, and effective batch size as HINT-WAM (\cref{tab:train_details}). The BC control is Stage B with the distillation term removed ($\lambda_{\mathrm{dist}}=0$, $\lambda_{\mathrm{BC}}=1$): it starts from SFT at step 3{,}000 and trains the action expert for 2{,}000 further steps with the flow-matching loss of \cref{eq:fm}, so it matches HINT-WAM in initialization, trainable parameters, and number of steps but has no teacher. Full SFT follows Stage 0 ($\lambda_{\mathrm{BC}}=2$, 5{,}000 steps from the released checkpoint), but trains both the video and the action expert (6.02B parameters); we report its final checkpoint. On RoboTwin, the BC control, full SFT, and HINT-WAM all start from the released RoboTwin checkpoint and train for 2{,}000 steps.

\begin{table*}[t]
    \centering
    \small
    \setlength{\tabcolsep}{6pt}
    \caption{\textbf{Training settings of each stage and baseline on LIBERO-90.} All runs share the optimizer, schedule, and effective batch size of 192.}
    \label{tab:train_details}
    \begin{tabular}{@{}lccccc@{}}
        \toprule
        & \multicolumn{3}{c}{HINT-WAM} & \multicolumn{2}{c}{Baselines} \\
        \cmidrule(r){2-4}\cmidrule(l){5-6}
        & Stage 0 (SFT) & Stage A (teacher) & Stage B & BC control & Full SFT \\
        \midrule
        Init.         & released & SFT@3k, frozen & SFT@3k & SFT@3k & released \\
        Trainable     & action expert & $\psi$ (enc.+adapt.) & action expert & action expert & video + action experts \\
        \#Params      & 1.02B & 13M & 1.02B & 1.02B & 6.02B \\
        Loss          & \cref{eq:fm}, $\lambda_{\mathrm{BC}}=2$ & \cref{eq:teacher} & \cref{eq:total} & \cref{eq:fm}, $\lambda_{\mathrm{BC}}=1$ & \cref{eq:fm}, $\lambda_{\mathrm{BC}}=2$ \\
        Learning rate & $10^{-4}$ & $3\times10^{-4}$ & $10^{-4}$ & $10^{-4}$ & $10^{-4}$ \\
        Steps         & 5{,}000 & 2{,}000 & 2{,}000 & 2{,}000 & 5{,}000 \\
        s / step      & 40.2 & 7.9 & 60.6 & 40.2 & 91.4 \\
        \bottomrule
    \end{tabular}
\end{table*}

\section{Training Cost}
\label{sec:supp_cost}

We measure the training time on 8 GPUs with 24\,GB of memory from checkpoint timestamps, so the reported times include periodic evaluation and checkpoint saving. GPU-hours are computed as steps $\times$ seconds per step $\times\,8/3600$. \cref{tab:supp_cost} lists the cost of every stage and pipeline on LIBERO-90.

\begin{table}[t]
    \centering
    \small
    \setlength{\tabcolsep}{3.5pt}
    \caption{\textbf{Training cost on LIBERO-90.} Top: individual stages. Bottom: complete pipelines and their LIBERO-90 success rates.}
    \label{tab:supp_cost}
    \begin{tabular}{@{}lcccc@{}}
        \toprule
        Stage & Steps & s/step & Hours & GPU-h \\
        \midrule
        SFT (action expert) & 3k / 5k & 40.2 & 33.5 / 55.8 & 268 / 447 \\
        Stage A (teacher)   & 2k & 7.9  & 4.4  & 35 \\
        Stage B (distill.)  & 2k & 60.6 & 33.7 & 269 \\
        BC control          & 2k & 40.2 & 22.3 & 179 \\
        Full SFT            & 5k & 91.4 & 127  & 1{,}016 \\
        \midrule
        Pipeline & & Hours & GPU-h & Success \\
        \midrule
        SFT 5k              & & 55.8 & 447 & 78.88 \\
        SFT 3k + BC 2k      & & 55.8 & 447 & 71.56 \\
        Full SFT            & & 127  & 1{,}016 & 81.24 \\
        \textbf{HINT-WAM}   & & \textbf{71.6} & \textbf{572} & \textbf{85.18} \\
        \bottomrule
    \end{tabular}
\end{table}

HINT-WAM, consisting of SFT for 3{,}000 steps followed by Stages A and B, costs $1.28\times$ the 5{,}000-step SFT pipeline and $0.56\times$ full SFT. A full-SFT step costs $2.3\times$ an action-expert step (91.4 vs.\ 40.2\,s) because the 5B-parameter video expert is also trained. A Stage-B step costs $1.51\times$ a BC step, since the student draws $M=4$ action chunks and the teacher one, all without gradients. Inference is unchanged: the deployed student is the unmodified Fast-WAM with 10 Euler steps, and the teacher and the hindsight encoder are discarded after training. On RoboTwin, Stage A takes 18.1\,s per step (80 GPU-hours for 2{,}000 steps), Stage B 60.6\,s (269 GPU-hours), and the BC control 37.8\,s (168 GPU-hours).

\section{Hindsight Teacher Checks}
\label{sec:supp_teacher}

Before Stage B, we verify that hindsight lowers the teacher's flow-matching loss on held-out frames, using 200 held-out batches and a paired comparison of the teacher with and without $\phi$. \cref{tab:supp_teacher} shows that hindsight reduces the action loss of every teacher, and that the advantage of the LIBERO-90 teacher keeps growing from 1{,}000 to 2{,}000 training steps.

\begin{table}[t]
    \centering
    \footnotesize
    \setlength{\tabcolsep}{4pt}
    \caption{\textbf{Hindsight teacher checks.} Relative change in the held-out action loss of the teacher with versus without hindsight $\phi$.}
    \label{tab:supp_teacher}
    \begin{tabular}{@{}lc@{}}
        \toprule
        Teacher & Loss change \\
        \midrule
        LIBERO-90, built on SFT                   & $-3.28\%$ ($t=5.77$) \\
        LIBERO-90 + replay, built on SFT + replay & $-3.06\%$ ($t=4.68$) \\
        RoboTwin 2.0, built on released checkpoint & $-5.37\%$ ($t=3.43$) \\
        LIBERO-90, 1{,}000 vs.\ 2{,}000 steps     & $-2.36\%\rightarrow-3.28\%$ \\
        \bottomrule
    \end{tabular}
\end{table}

\paragraph{The teacher must be built on the model being distilled.}
In a 20-task pilot with 10 trials per task, a teacher built on the weaker released checkpoint instead of the SFT model pulled the student down from $55.0\%$ to $47.5\%$, below SFT (with the velocity-matching objective). This supports the design choice in \cref{sec:method_teacher} that the teacher shares the student's initialization $\theta_0$.

\section{Full Paired Statistics on LIBERO-90}
\label{sec:supp_stats}

All tests are paired over the 90 LIBERO-90 tasks with 50 trials each. For every comparison, \cref{tab:supp_stats} reports the mean difference in success rate, its standard error, the paired $t$-statistic, the numbers of tasks on which the first method wins, loses, or ties, and the exact two-sided sign test over the non-tied tasks. The standard error for shuffled hindsight is derived from its mean difference and $t$-statistic. All LIBERO-90 numbers in the paper, including the released Fast-WAM and the BC control, use 50 trials per task. SFT reaches $78.88\%$ at both step 3{,}000 and step 5{,}000.

\begin{table*}[t]
    \centering
    \small
    \setlength{\tabcolsep}{8pt}
    \caption{\textbf{Paired statistics on LIBERO-90.} Differences are in \% over 90 tasks with 50 trials each.}
    \label{tab:supp_stats}
    \begin{tabular}{@{}lccccc@{}}
        \toprule
        Comparison & Mean diff. & SE & $t$ & Wins / losses / ties & Sign test $p$ \\
        \midrule
        HINT-WAM vs.\ SFT                    & $+6.30$ & 1.76 & 3.58 & 36 / 22 / 32 & 0.087 \\
        HINT-WAM vs.\ self-imitation         & $+5.56$ & 1.88 & 2.96 & 47 / 23 / 20 & 0.0056 \\
        HINT-WAM vs.\ shuffled hindsight     & $+4.90$ & 1.78 & 2.76 & 45 / 25 / 20 & 0.022 \\
        HINT-WAM vs.\ velocity matching      & $+6.16$ & 1.59 & 3.62 & 48 / 17 / 25 & 0.00042 \\
        Velocity matching vs.\ SFT           & $+0.14$ & 1.52 & 0.09 & 30 / 37 / 23 & 0.46 \\
        HINT-WAM + replay vs.\ SFT + replay  & $+4.86$ & --   & 2.66 & 44 / 23 / 23 & 0.014 \\
        \bottomrule
    \end{tabular}
\end{table*}

\section{Per-Task Analysis}
\label{sec:supp_pertask}

\begin{figure*}[t]
    \centering
    \includegraphics[width=\textwidth]{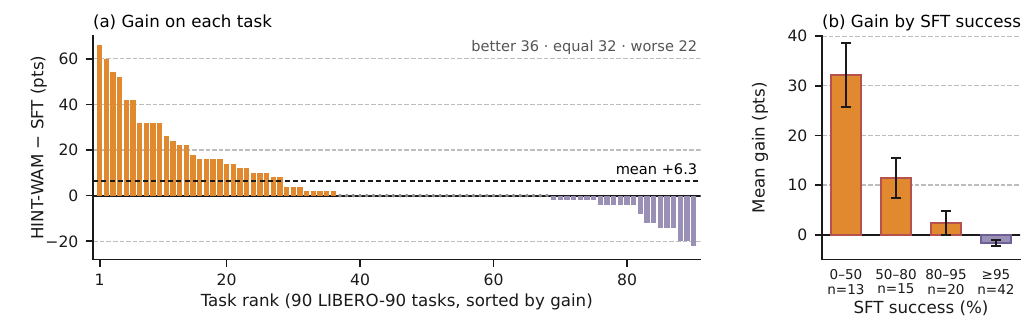}
    \caption{\textbf{Per-task gain of HINT-WAM over SFT on LIBERO-90.} (a)~Gain on each task, sorted. (b)~Mean gain ($\pm$ standard error) with tasks grouped by the success rate of SFT.}
    \label{fig:supp_pertask_gain}
\end{figure*}

\cref{fig:supp_pertask_gain} shows how the gain is distributed over the 90 tasks. HINT-WAM improves 36 tasks, leaves 32 unchanged, and lowers 22. The gain is concentrated on the tasks where SFT is weak: $+32.2\%$ on the 13 tasks where SFT succeeds less than half of the time and $+11.5\%$ on the 15 tasks between $50\%$ and $80\%$. On the 42 tasks that SFT already solves ($\geq95\%$), the change is small ($-1.7\%$, i.e., less than one failed episode out of 50). The many ties and small losses on saturated tasks explain why the sign test is weaker than the paired $t$-test (\cref{tab:supp_stats}). Since tasks are grouped by the success rate of SFT itself, part of this trend may reflect regression to the mean.

Although HINT-WAM improves the average success rate, it does not improve every task. \cref{tab:supp_pertask} lists the tasks with the largest drops for three pairs of methods. Most drops of HINT-WAM relative to SFT occur on pick-and-place tasks into a basket or a caddy compartment. Task 74 (``pick up the book and place it in the left compartment of the caddy'') is hurt by both distillation objectives, but much more by velocity matching ($80\%\rightarrow32\%$) than by the energy score ($80\%\rightarrow60\%$).

\begin{table}[t]
    \centering
    \small
    \setlength{\tabcolsep}{2.6pt}
    \caption{\textbf{Largest per-task drops on LIBERO-90.} Success rates (\%) over 50 trials per task.}
    \label{tab:supp_pertask}
    \begin{tabular}{@{}ccccccccc@{}}
        \toprule
        \multicolumn{3}{c}{HINT-WAM vs.\ SFT} & \multicolumn{3}{c}{MSE vs.\ SFT} & \multicolumn{3}{c}{HINT-WAM vs.\ MSE} \\
        \cmidrule(r){1-3}\cmidrule(lr){4-6}\cmidrule(l){7-9}
        Task & SFT & Ours & Task & SFT & MSE & Task & MSE & Ours \\
        \midrule
        52 & 80  & 58 & 74 & 80  & 32 & 76 & 86 & 60 \\
        53 & 60  & 40 & 89 & 90  & 44 & 53 & 52 & 30 \\
        74 & 80  & 60 & 86 & 60  & 28 & 75 & 48 & 30 \\
        5  & 100 & 86 & 58 & 80  & 52 & 21 & 96 & 80 \\
        78 & 100 & 86 & 6  & 100 & 74 & 30 & 94 & 80 \\
        \bottomrule
    \end{tabular}
\end{table}

Task instructions: 5, ``put the chocolate pudding in the top drawer of the cabinet and close it''; 52, ``pick up the milk and put it in the basket''; 53, ``pick up the orange juice and put it in the basket''; 74, ``pick up the book and place it in the left compartment of the caddy''; 78, ``pick up the book and place it in the front compartment of the caddy''.

\section{RoboTwin 2.0 Details}
\label{sec:supp_robotwin}

\paragraph{Hard subset.}
We keep the 25 tasks on which the released checkpoint scores below $95\%$ in either the clean or the randomized setting, measured with 20 episodes per task before any of our training: \texttt{blocks\_ranking\_size}, \texttt{handover\_block}, \texttt{move\_can\_pot}, \texttt{move\_stapler\_pad}, \texttt{hanging\_mug}, \texttt{open\_microwave}, \texttt{pick\_diverse\_bottles}, \texttt{place\_a2b\_left}, \texttt{place\_a2b\_right}, \texttt{place\_bread\_basket}, \texttt{place\_bread\_skillet}, \texttt{place\_can\_basket}, \texttt{place\_dual\_shoes}, \texttt{place\_mouse\_pad}, \texttt{place\_object\_basket}, \texttt{place\_object\_scale}, \texttt{place\_object\_stand}, \texttt{put\_bottles\_dustbin}, \texttt{put\_object\_cabinet}, \texttt{rotate\_qrcode}, \texttt{scan\_object}, \texttt{stack\_blocks\_three}, \texttt{stack\_bowls\_three}, \texttt{stamp\_seal}, and \texttt{turn\_switch}. Evaluation uses 50 episodes per task and setting. All methods start from the released checkpoint with the same data, 2{,}000 steps, a learning rate of $10^{-4}$, and a batch size of 192.

\paragraph{Paired tests.}
\cref{tab:supp_robotwin} reports paired tests over the 25 tasks. HINT-WAM improves over the BC control in both settings, with borderline significance in the clean setting, while its difference from the released checkpoint is not significant. Continued full SFT also lowers success, by $2.36\%$ in the clean and $0.20\%$ in the randomized setting, and HINT-WAM exceeds it by $2.92$ and $0.44\%$, respectively. Since the released checkpoint was already trained on this data, little headroom remains: continued training lowers success, whether by BC or by full fine-tuning, whereas HINT-WAM keeps it at or slightly above the level of the released checkpoint.

\begin{table*}[t]
    \centering
    \small
    \setlength{\tabcolsep}{8pt}
    \caption{\textbf{Paired tests on the RoboTwin 2.0 hard subset.} Mean difference in \% (SE, $t$) and wins / losses / ties over 25 tasks.}
    \label{tab:supp_robotwin}
    \begin{tabular}{@{}lcc@{}}
        \toprule
        Comparison & Clean & Randomized \\
        \midrule
        HINT-WAM vs.\ BC control   & $+1.68$ (0.84, $t=2.00$), 12 / 9 / 4 & $+1.60$ (1.32, $t=1.21$), 14 / 9 / 2 \\
        HINT-WAM vs.\ released     & $+0.56$ (1.19, $t=0.47$), 11 / 9 / 5 & $+0.24$ (1.04, $t=0.23$), 13 / 9 / 3 \\
        BC control vs.\ released   & $-1.12$ (1.08, $t=-1.04$), 9 / 11 / 5 & $-1.36$ (1.43, $t=-0.95$), 8 / 15 / 2 \\
        \bottomrule
    \end{tabular}
\end{table*}

\begin{figure*}[t]
    \centering
    \includegraphics[width=0.8\textwidth]{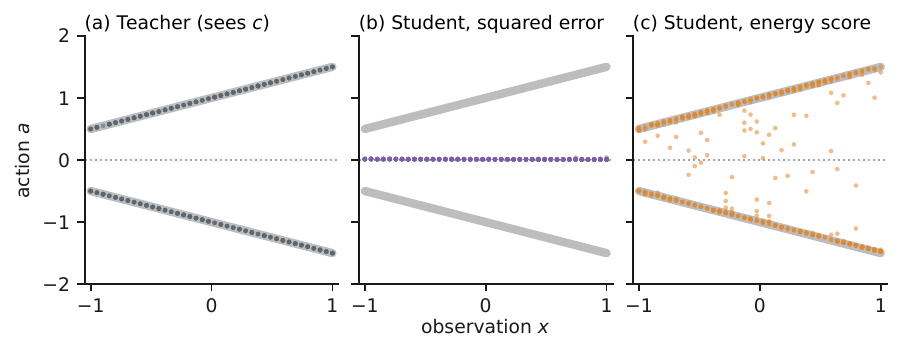}
    \caption{\textbf{Toy example of distillation under hidden information.} A hidden variable $c$ selects one of two correct actions; the teacher observes it, the student does not. (a)~Teacher samples. (b)~Squared-error distillation collapses to the mean of the two modes. (c)~The energy score recovers both.}
    \label{fig:toy}
\end{figure*}

\section{Evaluation Protocol on LIBERO-90}
\label{sec:supp_protocol}

Training and evaluation use the same 90 tasks but not the same episodes. Training uses the recorded human demonstrations (3{,}921 episodes, 569K frames), of which $10\%$ (393 episodes, split once with a fixed seed) are excluded from training and used only for offline loss metrics. Evaluation runs 50 new closed-loop episodes per task from the fixed initial states provided by LIBERO, which are generated separately from the demonstrations. This is the standard LIBERO protocol, also used by OpenVLA~\cite{kim2024openvla}, $\pi_0$~\cite{black2024pi0}, and Fast-WAM~\cite{yuan2026fastwam}. We find no sign that the test placements are memorized: the SFT checkpoint at step 3{,}000 reaches $80.94\%$ on 1{,}800 random placements, which never coincide with the test placements, and $78.88\%$ on the fixed test placements. Our evaluation thus measures generalization to held-out initial configurations of the training tasks rather than to unseen tasks.

\section{Self-Imitation Baseline}
\label{sec:supp_selfimit}

The SFT checkpoint at step 3{,}000 was rolled out from 20 random initial placements per task, none of which coincides with the test placements, yielding 1{,}800 episodes, of which 1{,}457 ($80.94\%$) succeeded. We added 1{,}454 successful episodes (203K frames, about $26\%$ of the training frames) to the LIBERO-90 demonstrations, excluding three episodes whose instruction does not appear in the training data. Training then followed Stage B of HINT-WAM with the behavior-cloning loss only, starting from SFT step 3{,}000 for 2{,}000 steps with a learning rate of $10^{-4}$ and a batch size of 192.

\section{Selection of the Objective and Learning Rate}
\label{sec:supp_selection}

Most hyperparameters are inherited from Fast-WAM or fixed by design, and the loss weights, adapter placement, and hindsight size were not tuned. The few design decisions we made were based on the held-out LIBERO-90 episodes and a 20-task pilot on LIBERO-90 with 10 trials per task; no tuning was performed on RoboTwin.
\begin{itemize}[leftmargin=*,itemsep=1pt,topsep=2pt]
    \item \textbf{Learning rate.} An initial round with a learning rate of $10^{-5}$ for 500 steps left every method under-trained, at or below $16\%$ on LIBERO-90. We therefore used $10^{-4}$, the fine-tuning learning rate of Fast-WAM.
    \item \textbf{Objective.} On the held-out episodes, the energy score reduced the error of the generated actions more than velocity matching ($23.7\%$ vs.\ $15.5\%$ relative to the baseline). The closed-loop results confirm this choice ($85.18\%$ vs.\ $79.02\%$, \cref{tab:supp_stats}).
    \item \textbf{Supervised points.} With the $\Delta_i^2$ weighting on our 10-step schedule, the last point carries $63.6\%$ of the mass and the last three points $91.1\%$. We therefore use $K=1$ for the energy score and $K=3$ for velocity matching. With $K=1$, the selected point is $\sigma_N$, so the implied action $\hat{\mathbf{a}}_{m,1}$ is exactly the output of the $m$-th student rollout.
\end{itemize}

\section{Toy Experiment: Mode Averaging}
\label{sec:supp_toy}

The toy example in \cref{fig:toy} isolates the mode-averaging effect of velocity matching. A hidden binary variable $c$ decides which of two actions is correct; the teacher observes it, and the student does not. Distilled with the squared error, the student predicts the average of the two actions, which lands on neither mode (success $0.00$). Distilled with the energy score from the same single teacher samples and $M=4$, it places its samples on both modes (success $0.92$). The same mechanism explains why velocity matching yields no gain on LIBERO-90, while the energy score improves success by $6.16\%$ over it (\cref{tab:supp_stats}).

\paragraph{Setup.}
The observation is $x\sim\mathcal{U}(-1,1)$, and a hidden $c\in\{-1,+1\}$, drawn independently of $x$ with equal probability, sets the correct action $a=c\,(1+0.5x)$, so the student cannot infer $c$ and no single action is correct for both values. The teacher observes $c$ and outputs the correct action, and each training state provides one teacher sample. The student is an implicit sampler $a=f(x,z)$ with $z\sim\mathcal{N}(0,1)$, implemented as an MLP with two hidden layers of 64 units and GELU activations. Both students are trained for 3{,}000 steps with Adam (learning rate $3\times10^{-3}$, batch size 256): the squared-error student regresses one sample onto the teacher sample, and the energy-score student matches $M=4$ samples to it. Success is the fraction of student samples within $0.25$ of either correct action, computed over 64 values of $x$ with 512 samples each and averaged over 3 seeds; a student that predicts the mean of the two actions therefore scores zero.

\section{Qualitative Rollouts}
\label{sec:supp_vis}

\begin{figure*}[p]
    \centering
    \begin{minipage}[t]{0.44\textwidth}\centering\includegraphics[width=\linewidth]{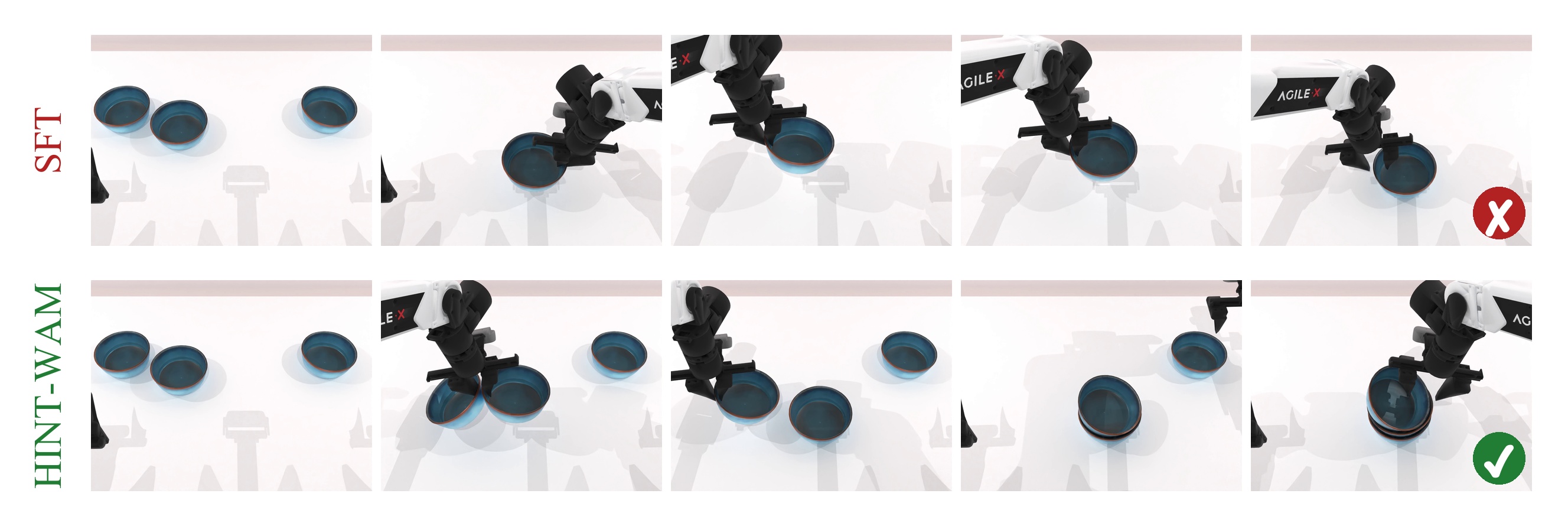}\\[-1pt]{\footnotesize (a) Episode 0}\end{minipage}\hfill
    \begin{minipage}[t]{0.44\textwidth}\centering\includegraphics[width=\linewidth]{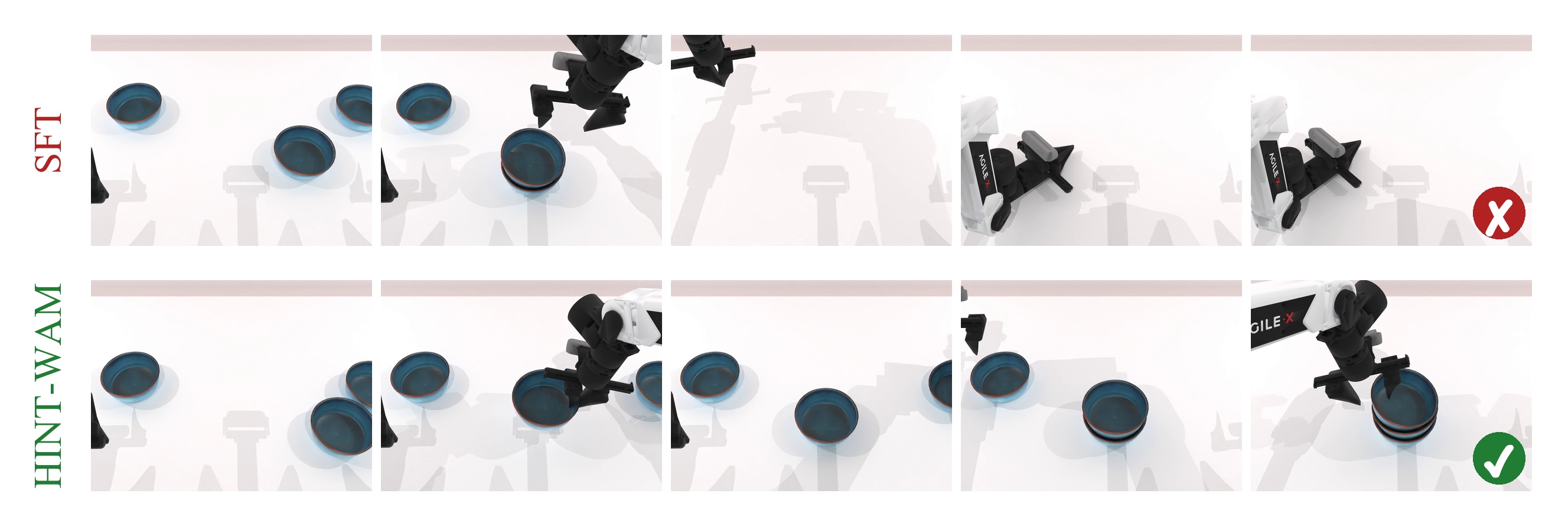}\\[-1pt]{\footnotesize (b) Episode 7}\end{minipage}\\[2pt]
    \begin{minipage}[t]{0.44\textwidth}\centering\includegraphics[width=\linewidth]{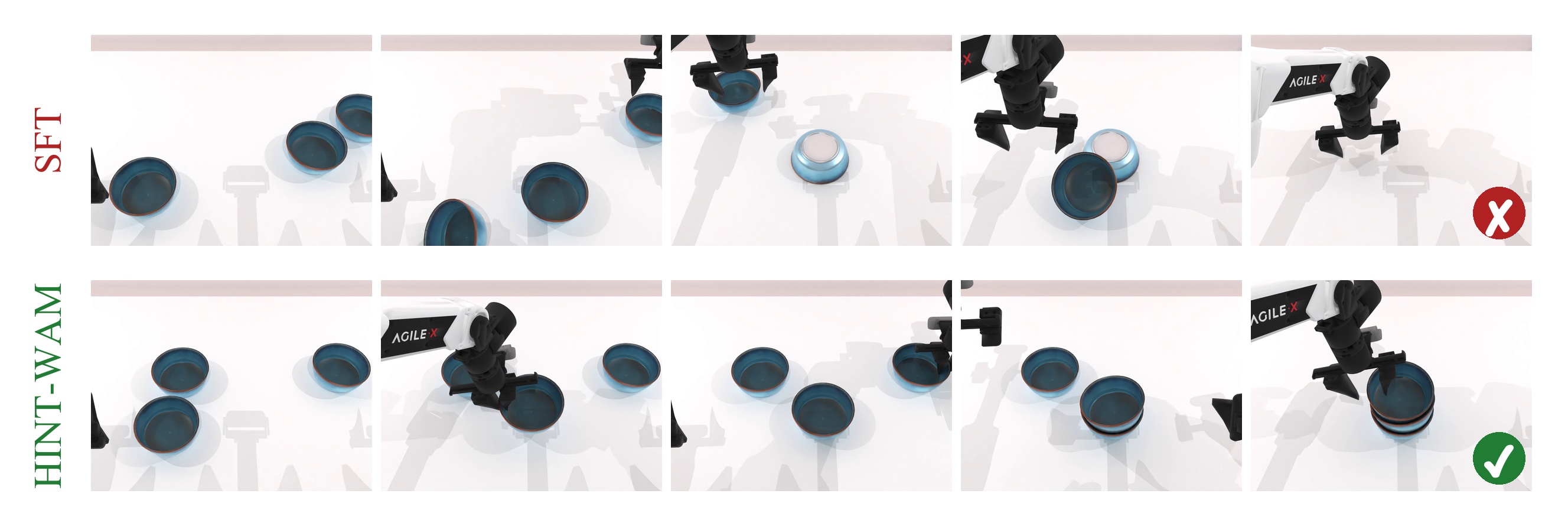}\\[-1pt]{\footnotesize (c) Episode 10}\end{minipage}\hfill
    \begin{minipage}[t]{0.44\textwidth}\centering\includegraphics[width=\linewidth]{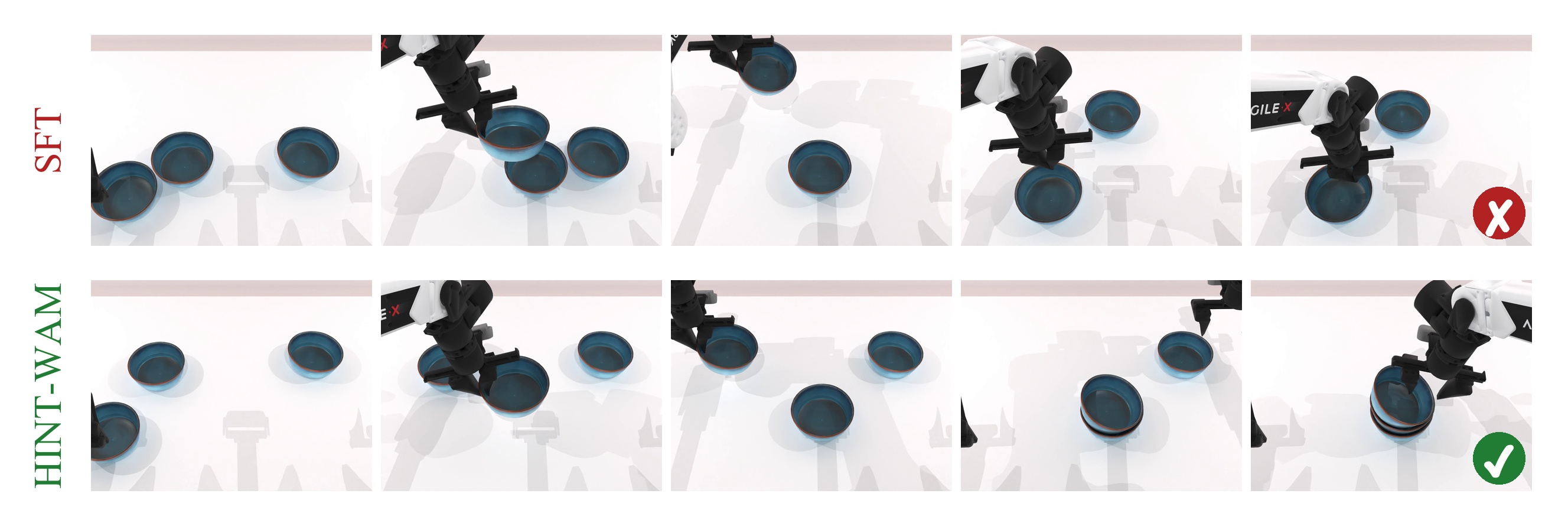}\\[-1pt]{\footnotesize (d) Episode 16}\end{minipage}
    \caption{\textbf{Rollouts on RoboTwin 2.0} (\texttt{stack\_bowls\_three}, clean setting). SFT (top row of each panel) fails, while HINT-WAM (bottom row) stacks all three bowls from the same initial state.}
    \label{fig:supp_vis_robotwin}

    \vspace{2pt}
    \begin{minipage}[t]{0.44\textwidth}\centering\includegraphics[width=\linewidth]{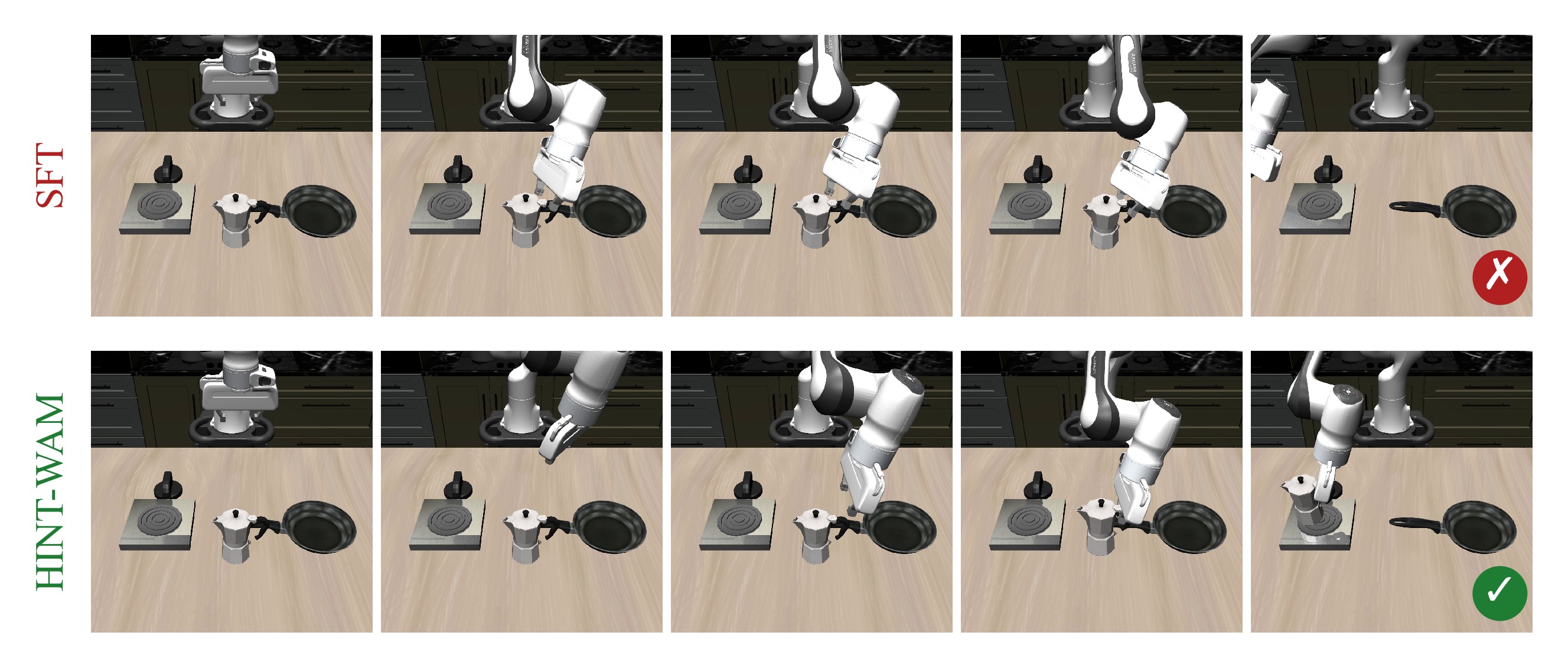}\\[-1pt]{\footnotesize (a) Task 19: ``put the moka pot on the stove''}\end{minipage}\hfill
    \begin{minipage}[t]{0.44\textwidth}\centering\includegraphics[width=\linewidth]{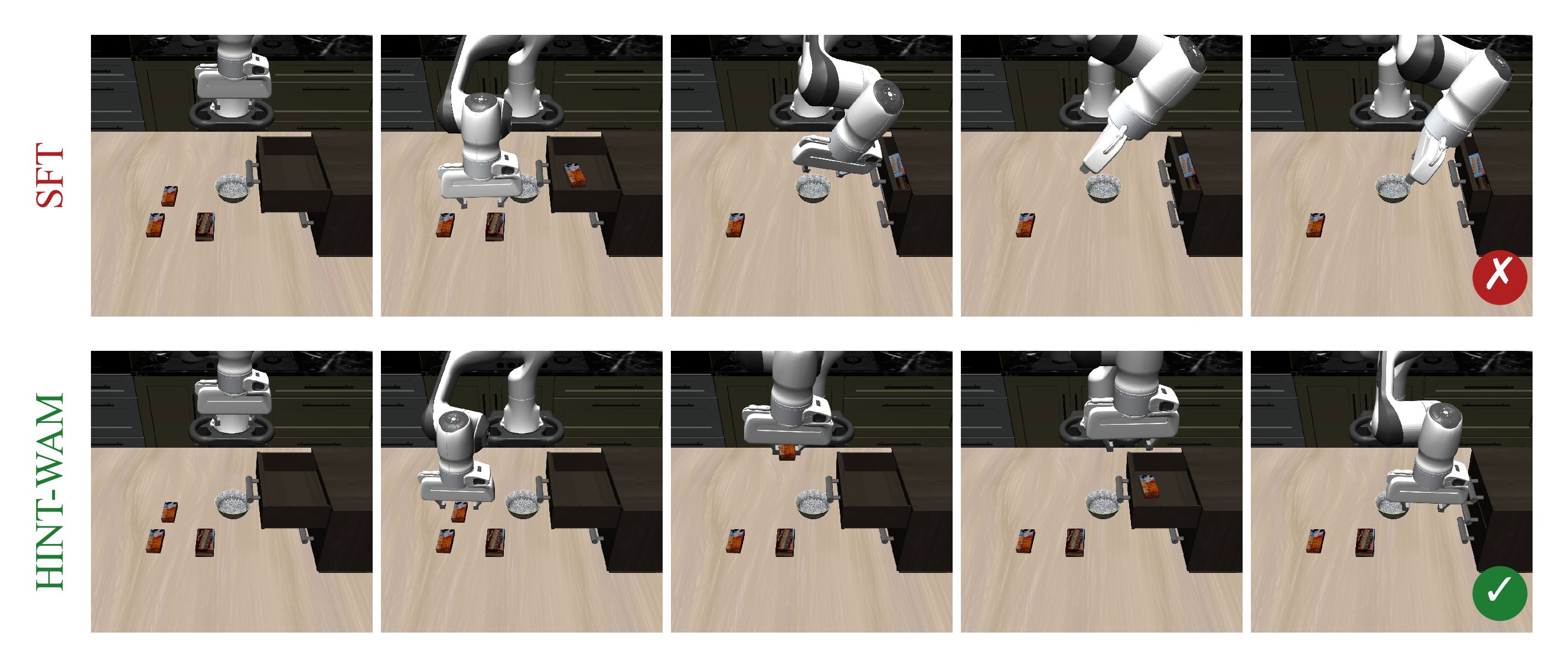}\\[-1pt]{\footnotesize (b) Task 3: ``put the butter at the back in the top drawer of the cabinet and close it''}\end{minipage}\\[2pt]
    \begin{minipage}[t]{0.44\textwidth}\centering\includegraphics[width=\linewidth]{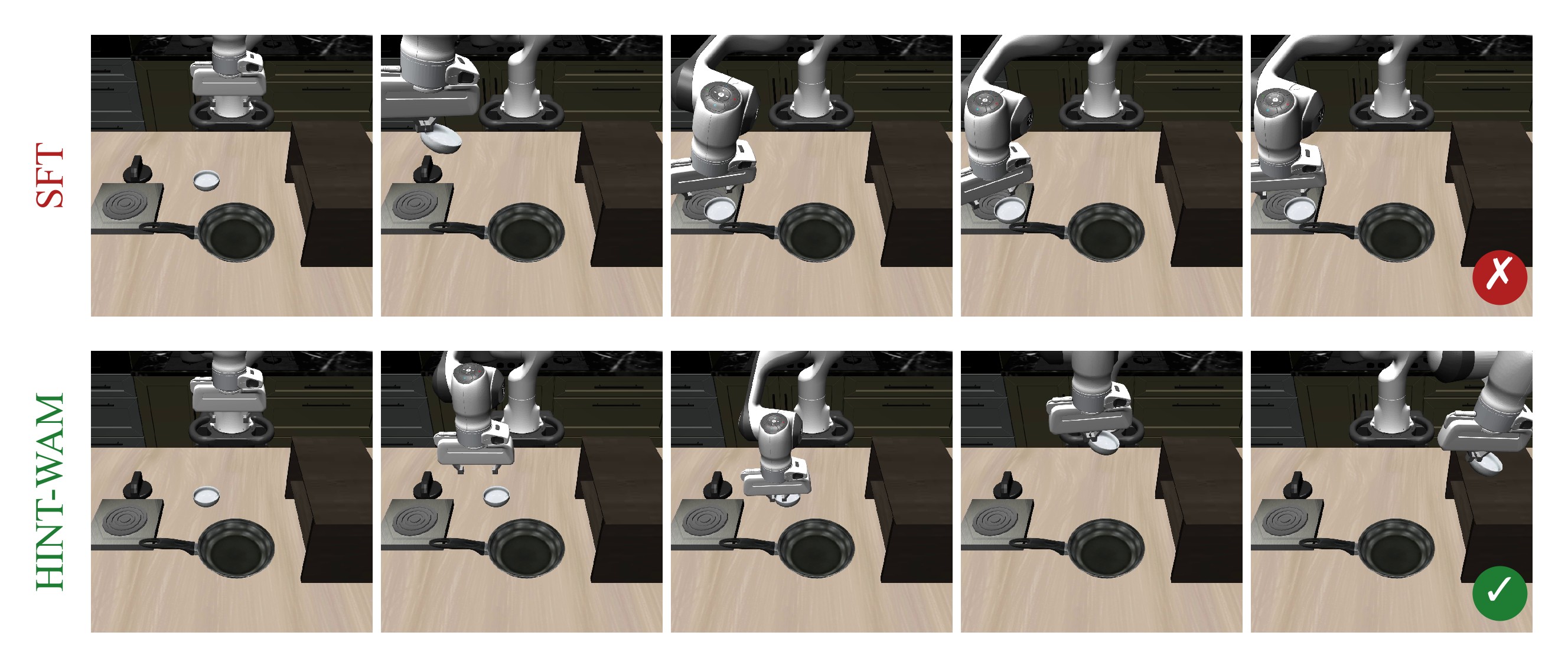}\\[-1pt]{\footnotesize (c) Task 43: ``put the white bowl on top of the cabinet''}\end{minipage}\hfill
    \begin{minipage}[t]{0.44\textwidth}\centering\includegraphics[width=\linewidth]{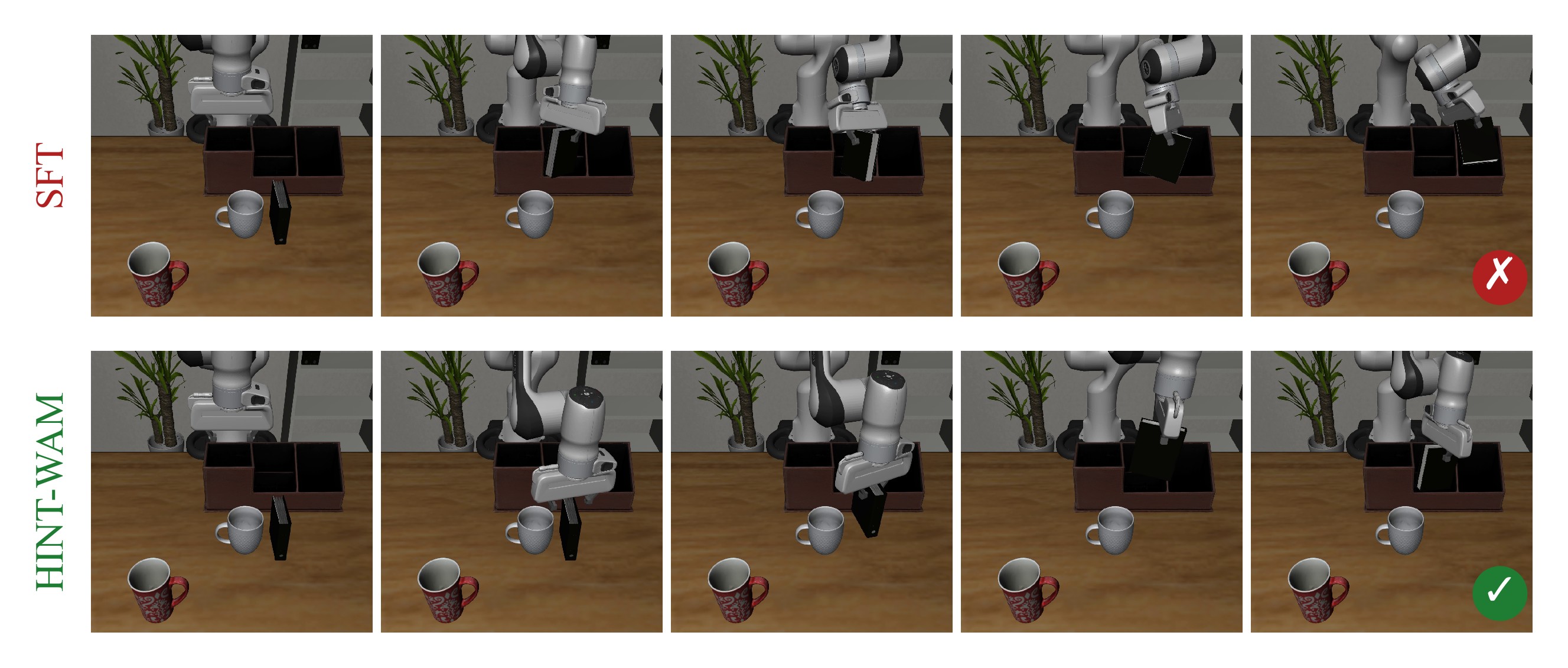}\\[-1pt]{\footnotesize (d) Task 81: ``pick up the book and place it in the front compartment of the caddy''}\end{minipage}
    \caption{\textbf{Rollouts on new LIBERO-90 tasks.} SFT (top row of each panel) fails, while HINT-WAM (bottom row) succeeds from the same initial state.}
    \label{fig:supp_vis_new}

    \vspace{2pt}
    \begin{minipage}[t]{0.325\textwidth}\centering\includegraphics[width=\linewidth]{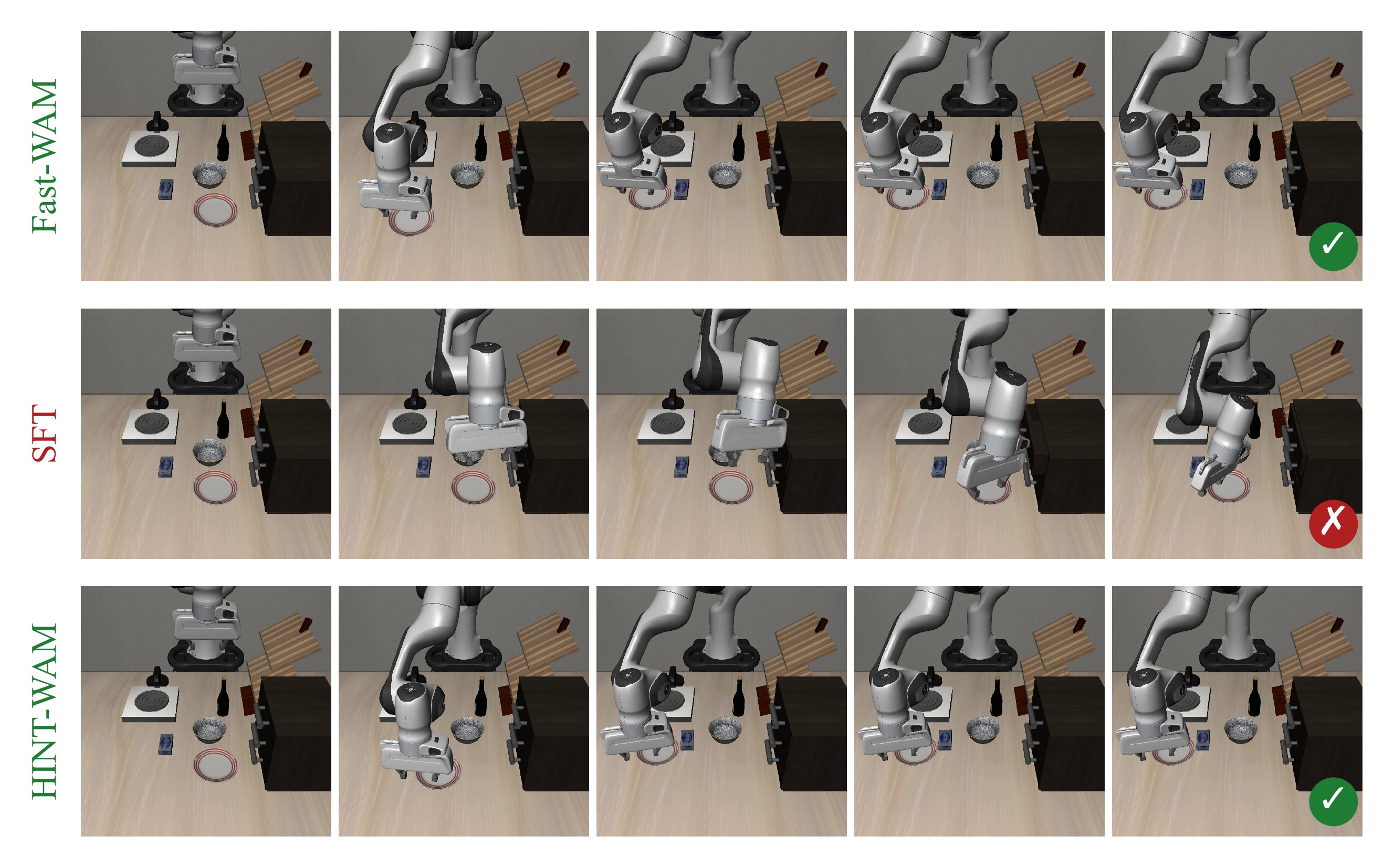}\\[-1pt]{\footnotesize (a) LIBERO-Goal: ``push the plate to the front of the stove''}\end{minipage}\hfill
    \begin{minipage}[t]{0.325\textwidth}\centering\includegraphics[width=\linewidth]{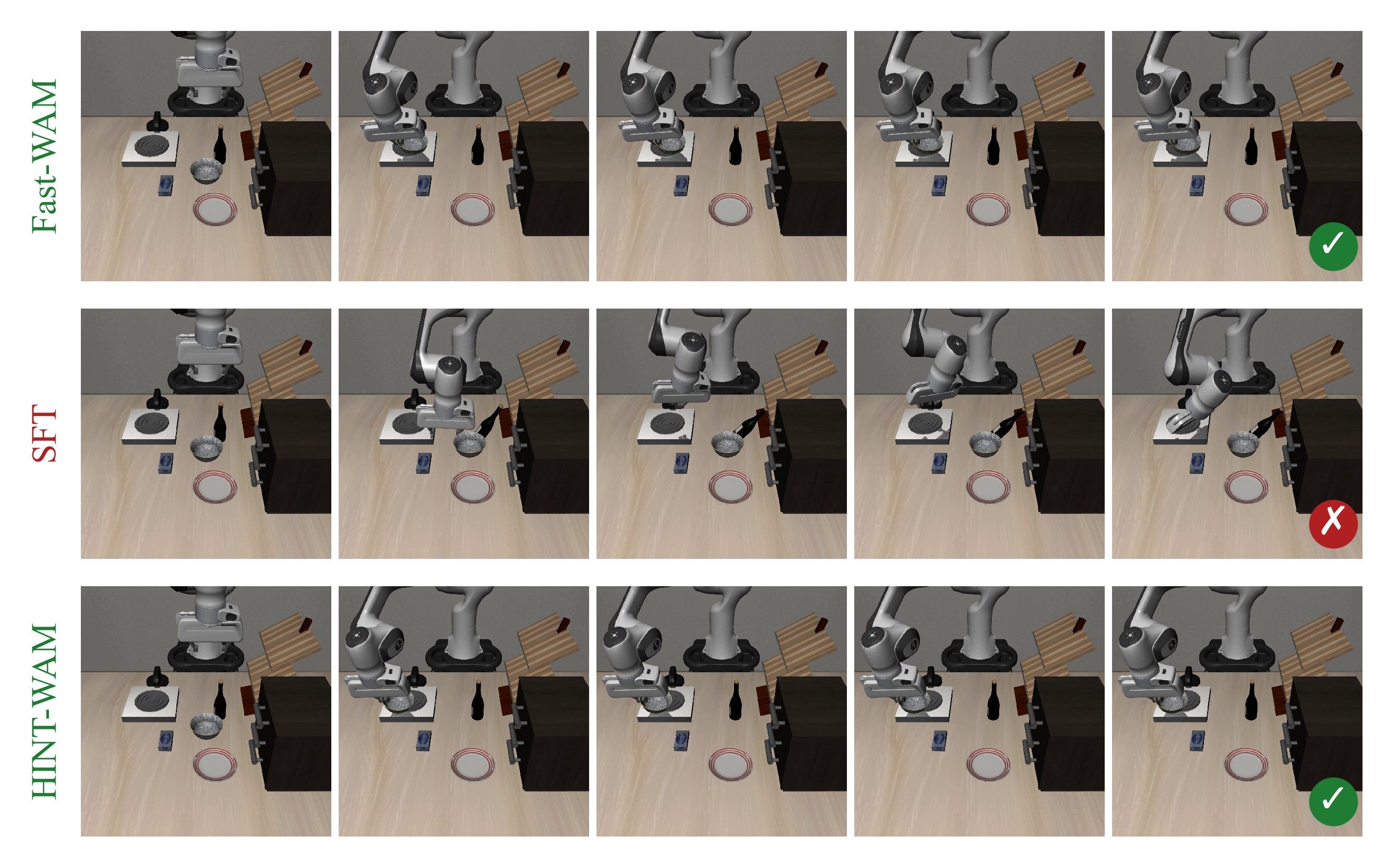}\\[-1pt]{\footnotesize (b) LIBERO-Goal: ``put the bowl on the stove''}\end{minipage}\hfill
    \begin{minipage}[t]{0.325\textwidth}\centering\includegraphics[width=\linewidth]{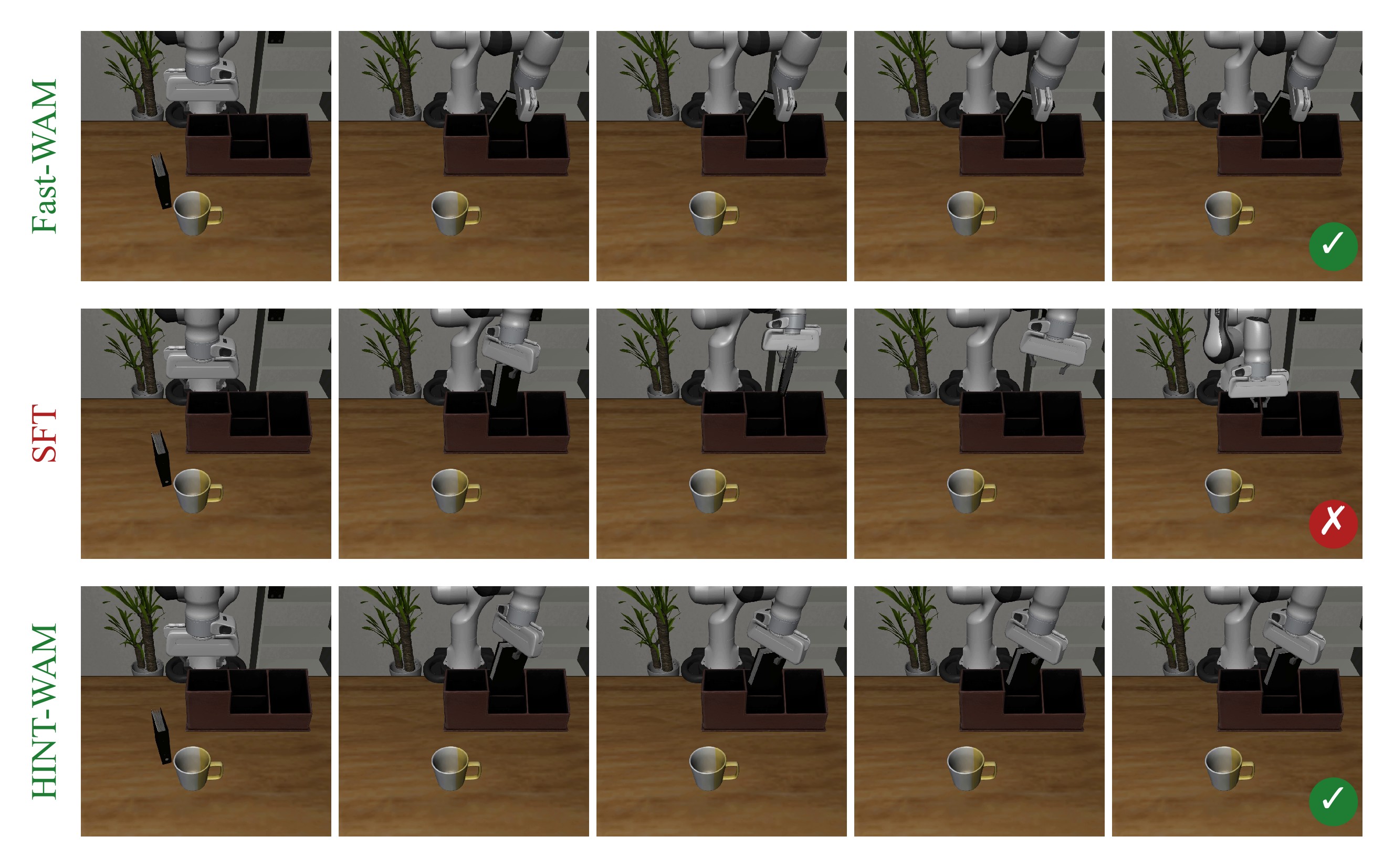}\\[-1pt]{\footnotesize (c) LIBERO-Long: ``pick up the book and place it in the back compartment of the caddy''}\end{minipage}
    \caption{\textbf{Rollouts on retained skills after post-training on LIBERO-90 without replay.} The released Fast-WAM (top) and HINT-WAM (bottom) succeed, while SFT (middle) fails, from the same initial state.}
    \label{fig:supp_vis_retained}
\end{figure*}

\cref{fig:supp_vis_new,fig:supp_vis_retained,fig:supp_vis_robotwin} show closed-loop rollouts in which HINT-WAM succeeds where the baseline fails. Each pair of rows starts from the same initial state; the frames are evenly spaced over the episode, and the badge on the last frame marks success or failure.

\paragraph{New tasks.}
\cref{fig:supp_vis_new} shows four LIBERO-90 tasks. The failures of SFT are typically errors in the final stage of the task rather than in grasping: the moka pot is not set down on the stove (task 19), the drawer is left open after the butter has been placed (task 3), the bowl is not placed on the cabinet (task 43), and the book is dropped into the wrong compartment of the caddy (task 81). HINT-WAM completes each of these tasks from the same initial state.

\paragraph{Retained skills.}
\cref{fig:supp_vis_retained} shows tasks from the original LIBERO suites, which the released Fast-WAM solves before post-training. After post-training on LIBERO-90 without replay, SFT fails these tasks, whereas HINT-WAM still solves them, consistent with its smaller forgetting in \cref{tab:cl}.

\paragraph{RoboTwin 2.0.}
\cref{fig:supp_vis_robotwin} shows four episodes of \texttt{stack\_bowls\_three} (clean setting). SFT moves single bowls but never completes the stack: it carries a bowl without placing it (a), loses track of the remaining bowls (b), flips a bowl (c), or stops after one lift (d), whereas HINT-WAM stacks all three. Both reach the same success rate on this task, so these episodes illustrate failure modes rather than a per-task gain.

\section{Discussion}
\label{sec:supp_discussion}

\paragraph{Why does a teacher that sees the future help?}
The student never receives $\phi$; the teacher supplies one action sample per state. Since the energy score is strictly proper, the student is driven towards the distribution of teacher actions given what it observes, \ie, the continuations consistent with the demonstrations, not one particular future. Hindsight thus resolves the ambiguity of the SFT target rather than giving information the student must recover. The gain requires the correct future: with the future of another episode, the same pipeline reaches only $80.28\%$ (\cref{tab:main}).

\paragraph{In what sense is HINT-WAM on-policy?}
The states are on-policy with respect to the action sampler: the student is supervised on the denoising states its own ODE visits; noised demonstrations instead reach only $80.74\%$ (\cref{fig:onpolicy}). The observations still come from the demonstrations, so HINT-WAM needs neither rewards nor a simulator, but does not address compounding errors across control steps and is complementary to DAgger or reinforcement learning.

\paragraph{Is the gain due to extra compute or tuning?}
No. The same 2{,}000 extra steps without the teacher lower success to $71.56\%$, and HINT-WAM exceeds full SFT by $3.94\%$ at $0.56\times$ its cost (\cref{tab:supp_cost}). Neither its loss weights nor its modules were tuned, and all hyperparameters are reused unchanged on RoboTwin.

